\documentclass{bmvc2k}
\usepackage{multirow}
\usepackage{makecell}
\usepackage{amssymb}
\usepackage{comment}

\title{
Robust Ellipsoid-specific Fitting via Expectation Maximization
}

\addauthor{\scriptsize{Mingyang Zhao}}{zhaomingyang16@mails.ucas.ac.cn}{1}
\addauthor{\scriptsize{Xiaohong Jia$^\dagger$}}{xhjia@amss.ac.cn}{2}
\addauthor{\scriptsize{Lei Ma$^\dagger$}}{lei.ma@pku.edu.cn}{4}
\addauthor{\scriptsize{Xinlin Qiu}}{qiuxinling@hbc.edu.cn}{5}
\addauthor{\scriptsize{Xin Jiang}}{jiangxin@buaa.edu.cn}{6}
\addauthor{\scriptsize{Dong-Ming Yan}}{yandongming@gmail.com}{3}
\addinstitution{
\scriptsize{Beijing Academy of Artificial Intelligence (BAAI) and NLPR, Institute of Automation, CAS, Beijing, China}
}
\addinstitution{
\scriptsize{Academy of Mathematics and Systems Science, CAS, Beijing, China}
}
\addinstitution{
\scriptsize{NLPR, Institute of Automation, CAS and the University of CAS, Beijing, China \\
\scriptsize{$^\dagger$ Corresponding author} }
}
\addinstitution{
\scriptsize{National Engineering Laboratory for Video Technology, Peking University, and BAAI, Beijing, China}
}
\addinstitution{
\scriptsize{College of Artificial Intelligence, Hubei Business College, Wuhan, Hubei}
}
\addinstitution{
\scriptsize{School of Mathematical Science, Beihang University, Beijing, China} 
}

\runninghead{\scriptsize{Zhao \etal}}{\scriptsize{Robust Ellipsoid-specific Fitting via Expectation Maximization}}

\def\etal{\emph{et al}\bmvaOneDot}
\def\ie{\emph{ie}\bmvaOneDot}

\usepackage{times}
\usepackage{graphicx}
\usepackage{amsmath}
\usepackage{amssymb}

\usepackage{subfigure}
\usepackage{blkarray}
\usepackage{threeparttable}
\usepackage{multirow}
\usepackage{makecell}
\usepackage{colortbl}
\usepackage{array}
\usepackage{diagbox}
\usepackage{booktabs}
\usepackage{tikz}
\usepackage{fbox}
\usepackage{bm}
\usepackage{bbm}
\usepackage{algpseudocode}
\usepackage{algorithm}
\usepackage{algorithmicx}
\usepackage{comment}
\usepackage{marginnote}

\newcommand{\revise}[1]{{\color{black}{#1}}}
\newcommand{\noin}[1]{{\noindent{\textbf{#1}}}}

\newcommand{\tr}{{\mathop{\mathrm{tr}}}}
\newcommand{\atan}{{\mathop{\mathrm{atan}}}}

\newcommand{\RDOS}{{\mathop{\mathrm{RDOS}}}}

\newcommand{\dist}{{\mathop{\mathrm{dist}}}}
\newcommand{\diag}{{\mathop{\mathrm{diag}}}}

\newtheorem{lemma}{Lemma}
\newtheorem{definition}{Definition}

\begin{document}

\maketitle

\begin{abstract}
	Ellipsoid fitting \revise{is} of general \revise{interest} in machine vision, such as object detection and shape approximation. Most existing approaches rely on the least-squares fitting of quadrics, minimizing the algebraic or geometric distances, with additional constraints to enforce the quadric as an ellipsoid. However, they are susceptible to outliers and  non-ellipsoid or biased results when the axis ratio exceeds certain thresholds.

	To address these problems, we propose a novel and robust method for ellipsoid fitting in a noisy, \revise{outlier-contaminated} 3D environment. We explicitly model the ellipsoid by \emph{kernel density estimation} (KDE)  of the input data. The ellipsoid fitting is cast as a \emph{maximum likelihood estimation} (MLE) problem without extra constraints, where a weighting term is added to depress outliers, and then effectively solved via the \emph{Expectation-Maximization (EM)} framework. Furthermore, we introduce the \emph{vector $\varepsilon$ technique} to accelerate the convergence of the original EM. The proposed method is compared with representative state-of-the-art approaches by extensive experiments, and results show that our method is ellipsoid-specific, parameter free, and more robust against noise, outliers, and the large axis ratio. Our implementation is available at \url{https://zikai1.github.io/}.
	
%
\end{abstract}

\section{Introduction}
\label{sec:intro}
Detecting and fitting quadratic surfaces or quadrics from 3D scattered points, such as planes, cylinders, and ellipsoids is a fundamental problem in machine vision~\cite{miller1988analysis,Faber01abuyer's,bischoff2002ellipsoid,blane20003l,allaire2007type,georgiev2016real,beale2016fitting}. Among quadrics, ellipsoids attract more \revise{interest} because they are the \revise{uniquely} bounded and centric surface, which provides a good characterization or approximation for the center and orientation of objects~\cite{tasdizen2001robust,li2004least,Iasonefficient11}. For instance, Rimon \etal~\cite{rimon1997obstacle} use ellipsoid fitting to approximate the robot shape and speed up the collision detection process.  Jia~\etal~\cite{jia2011algebraic} take ellipsoids as bounding box for continuous collision detection. Gietzelt \etal \cite{gietzelt2013performance} reduce the accelerometer calibration as a 3D ellipsoid fitting problem, by which the transformation and correction matrix is identified.
\begin{figure}
	\centering
	\subfigure[\revise{Outlier-contaminated} fitting]{
		\includegraphics[width=0.24\textwidth]{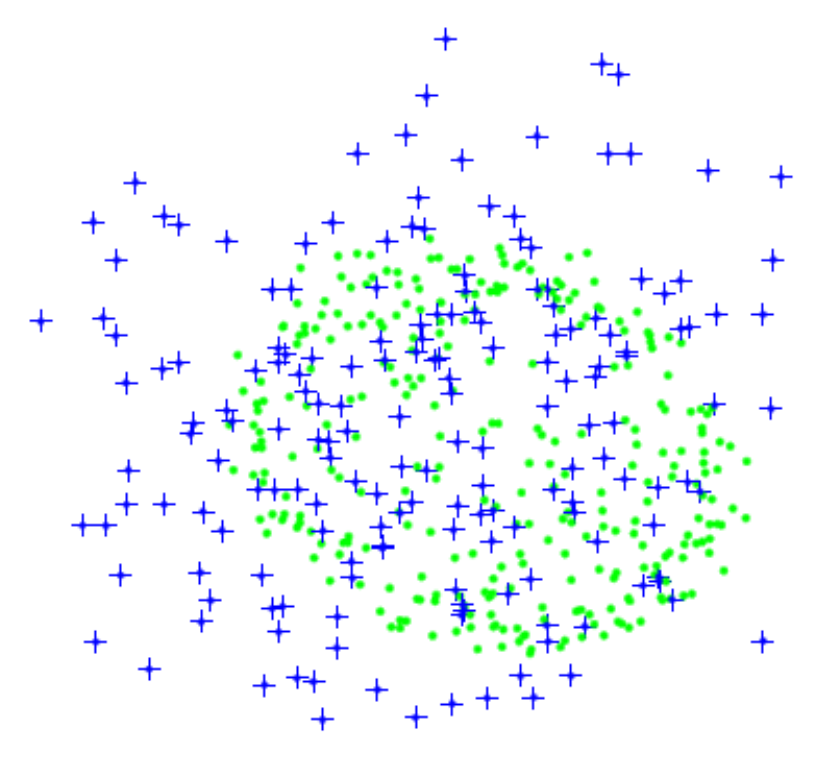}
		\includegraphics[width=0.24\textwidth]{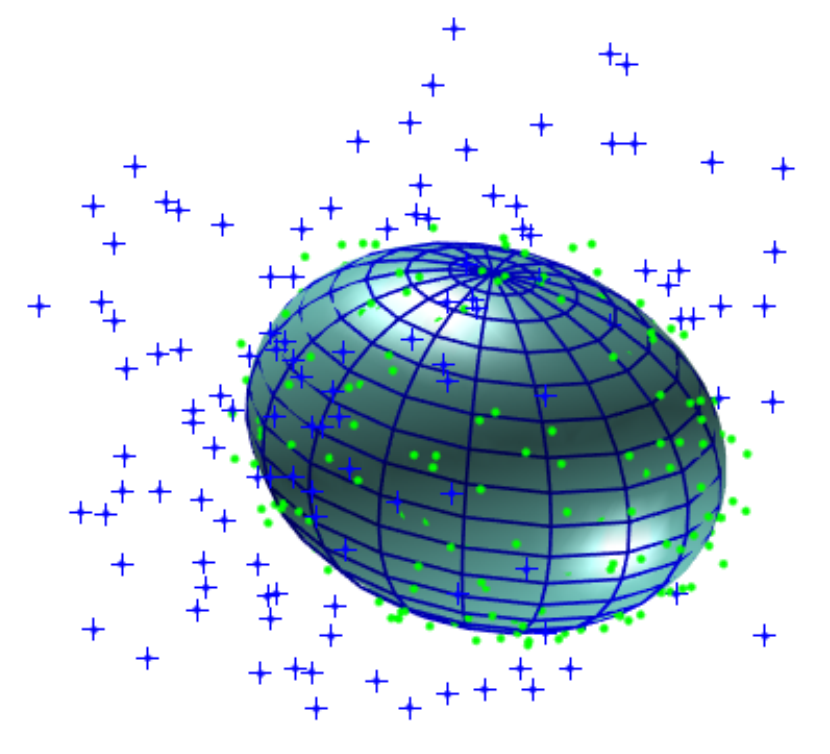}
}
\subfigure[3D medical data]{
	\includegraphics[width=0.19\textwidth]{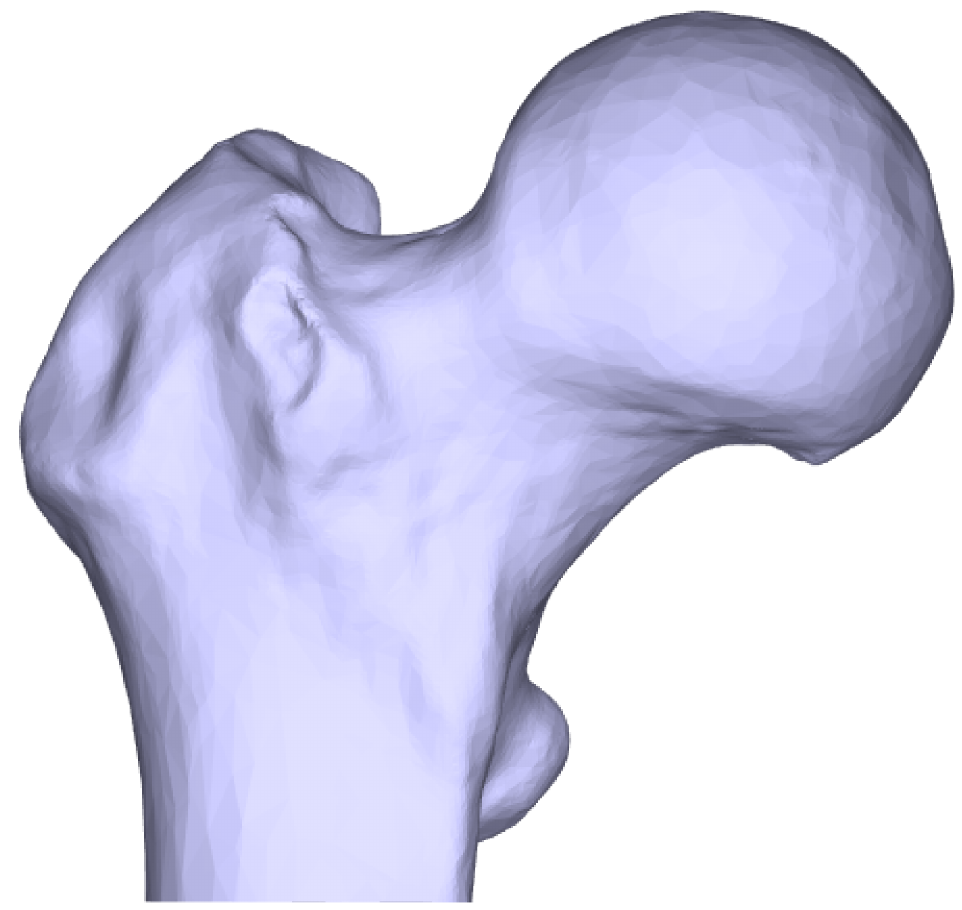}
	\includegraphics[width=0.21\textwidth]{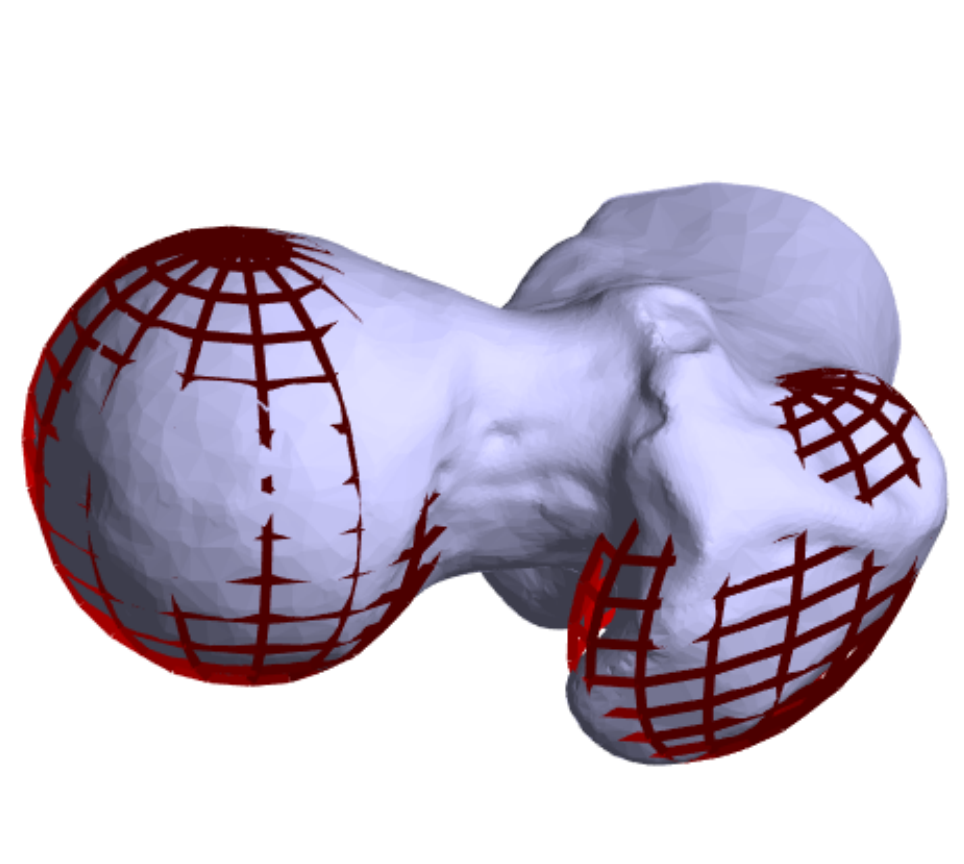}
}	
	\caption{Our method shows (a) highly accurate fitting in the contamination of heavy outliers (‘\revise{+}’) (net ellipsoid is the ground truth), and (b) approximation for 3D medical femur images.
}
	\vskip -0.5cm
	\label{fig:ex1}
\end{figure}
Most existing methods adopt the \emph{least-squares (LS)} principle for ellipsoid fitting, among which algebraic or geometric distances are minimized. These methods attain satisfactory results for simple and low-noise data points but are susceptible to outliers that are quite common and inevitable in practice~\cite{birdal2019generic,thurnhofer2020ellipse,zhao2021robust}. Meanwhile, various constraints have been investigated to force the fitted surface as an ellipsoid regardless of the input data. However, they cannot guarantee the best fitting \revise{when} the ratio between the longest axis and the shortest one surpasses certain thresholds, such as two in~\cite{li2004least} and~\cite{kesaniemi2017direct}, thereby significantly limiting their applications.

To overcome the shortcomings above, we propose a novel ellipsoid fitting method that dose not relying on LS, instead, \revise{by using a set of points sampled over a unit sphere and transformed by the model parameters}, which is highly robust against outliers, and is ellipsoid-specific regardless of the axis ratio. Inspired by a study of the point set registration framework in~\cite{myronenko2010point}, we explicitly model the ellipsoid and represent it via \textit{Gaussian mixture models} (GMM), armed with an adaptive uniform distribution to depress outliers. Then ellipsoid fitting is formulated as \revise{an} MLE without extra constraints, which is effectively solved by the \emph{expectation-maximization} (EM) framework. Furthermore, we encapsulate all parameters into a sequence and introduce the vector $\varepsilon$ algorithm~\cite{wang2008acceleration} to accelerate the EM convergence.

Our method is robust enough against outliers up to 60\% and is without handcraft tuning of hyper-parameters. The performance of our method regarding the accuracy and robustness is validated by measuring the offset and shape deviations on various numerical experiments. We further demonstrate the promising applications of the proposed method on real-world scanned point clouds, where occlusion and outliers exist. Furthermore, our method can be directly generalized to fit other quadrics such as cylinders and cones, \revise{as long as} a parametric representation is given. To summarize, the contributions of this work are threefold as follows:
\begin{itemize}
	\item A novel ellipsoid-specific fitting method with remarkable robustness against outliers, noise and the axis ratio.	
	\item The probabilistic method is applied for ellipsoid fitting. We explicitly model the ellipsoid based on the outlier analysis from the kernel density estimation and effectively speed up the convergence of the EM framework.
	\item All parameters are updated automatically by the derivation of the analytical gradients without user tuning.
\end{itemize}

\section{Related Work}
\begin{definition}
	A general quadric in 3D Euclidean space is defined by the zero set of a second order polynomial:
	\begin{eqnarray}
		\begin{aligned}
			Q(\mathbf{a},\mathbf{p})=\mathbf{a}\cdot\mathbf{p}&=Ax^2+By^2+Cz^2+2Dxy+2Exz\\
			&+2Fyz+2Gx+2Hy+2Iz+J=0,
		\end{aligned}
		\label{eq:quadric}
	\end{eqnarray}
	where $\mathbf{a}=[x^2\quad y^2\quad z^2\quad 2xy\quad 2xz\quad 2yz\quad 2x\quad 2y\quad 2z\quad 1]^T$ are built from the point $\mathbf{x}=(x, y, z)^T\in \mathbb{R}^3$, and $\mathbf{p}=[A\quad B\quad C\quad D\quad E\quad F\quad G\quad H\quad I\quad J]^T$ are the coefficients that characterize the quadric. 
\end{definition}
Eq.~\ref{eq:quadric} represents an ellipsoid if its quadratic invariants satisfy~\cite{harris1998handbook} 
\begin{eqnarray}
	I_1>0, \quad I_2*I_3>0,
\end{eqnarray}
\revise{where $I_1=AB+AC+BC-D^2-E^2-F^2$, $I_2=A+B+C$, and
\begin{eqnarray}
		I_3=\left|
		\begin{array}{ccc}
			A &D &E\\
			D &B &F\\
			F &E &C\\
		\end{array}
		\right|.
\end{eqnarray}}Given a set of data points $\mathbf{X}=\{\mathbf{x}_i\in \mathbb{R}^3\}_{i=1}^{N}$ that are sampled from a potential ellipsoid possibly with noise or outliers, our purpose is to fit an ellipsoid from the data. The most frequently used methods are those based on the LS principle, which can be classified into algebraic and geometric fittings. \\

\noindent{\textbf{Algebraic fitting.}} To find the optimal parameter $\mathbf{p}$, algebraic fitting 
minimizes the deviation of the polynomial in Eq.~\ref{eq:quadric} (\ie, the \emph{algebraic distance or equation error})~\cite{fitzgibbon95a} by
\begin{eqnarray}
	\sum_{i=1}^{N}Q^2(\mathbf{a}_i,\mathbf{p})=\sum_{i=1}^{N}(\mathbf{p}^T\mathbf{a}_i)^2=\mathbf{p}^T\mathbf{Q}\mathbf{p},
	\label{eq:algebraic}
\end{eqnarray}where $\mathbf{a}_i=\mathbf{a}(\mathbf{x}_i)$ is the vector corresponding to the $i^{th}$ point $\mathbf{x}_i=(x_i,  y_i, z_i)^{T}$, and $\mathbf{Q}=\sum_{i=1}^{N}\mathbf{a}_i\mathbf{a}_i^T\in \mathbb{S}^{10}$ is the \emph{scatter matrix}. 
Ellipse-specific fitting in 2D is solved by Fitzgibbon \etal~\cite{fitzgibbon1999direct}, and a direct extension for 3D ellipsoid-specific case is presented in~\cite{li2004least} under the determinant $(A-B-C)^2-4(F^2+G^2+H^2+BC)>0.$
Nevertheless, it attains a best fit only when the shortest axis of the ellipsoid is at least half of the longest one. Once this hypothesis fails, a bisection search must be executed to provide an approximation. Thus it may deviate from the ground truths. Recently, Kes{\"a}niemi \etal~\cite{kesaniemi2017direct} elaborate previous approaches and simultaneously consider three trace constraints $D^2-AB<0, E^2-AC<0, $ and $F^2-BC<0$, to force the quadric to be an ellipsoid, but it only credibly fits ellipsoids with \emph{a prior} that their maximal axis ratio $r_{ax}< \sqrt{\frac{2d-2}{d-2}}$, where $d$ is the dimension. When $d=3$ in our case, the limit value $r_{ax}=2$, meaning that it may fail to fit an ellipsoid whose longest axis is more than twice the shortest one. Therefore, similar to~\cite{li2004least}, the application scope of~\cite{kesaniemi2017direct} is also greatly confined. Furthermore, according to the \emph{Gauss-Markov theorem}~\cite{rousseeuw2005robust}, LS fitting is susceptible to outliers that are quite common in practice.

Several methods~\cite{calafiore2002approximation,ying2012fast} treat ellipsoid fitting as a \emph{semi-definite programming} (SDP) problem, where ellipsoid-specificity is formalized as the matrix semi-definiteness such that $S_{p}(\mathbf{A})\succeq0$, where $S_{p}(\cdot)$ is the operator that extracts the leading $p \times p$ \revise{principal} submatrix of $\mathbf{A}$. Lin \etal~\cite{lin2015fast} introduce \emph{alternating direction method of multipliers} (ADMM) to speed up SDP solving but still minimize the residual error $\|\mathbf{a}_i^T\mathbf{p}\|_2$ in the LS sense, thereby their method is sensitive to outlier-contaminated environment.\\

\noindent{\textbf{Geometric fitting.}}~Alternatively, geometric fitting~\cite{gander1994least,ahn2002orthogonal} minimizes the \emph{orthogonal distance} from point $\mathbf{x}_i, i=1, \cdots, N$, to the ellipsoid
\begin{eqnarray}
	\dist(\mathbf{X}, Q)=\sum_{i=1}^{N}\|\mathbf{x}_i-\bar{\mathbf{x}}_i\|^2=\sum_{i=1}^{N}d_{i}^2,
	\label{eq:gf}
\end{eqnarray}
where $\bar{\mathbf{x}}_i$ is the point on the ellipsoid closet to $\mathbf{x}_i$, and $\|\mathbf{x}-\bar{\mathbf{x}}_i\|$ denotes the Euclidean distance between $\mathbf{x}$ and $\mathbf{x}_i$. Geometric fitting exhibits more sound physical interpretations and higher accuracy than algebraic fitting, but it requires much more time for distance evaluation. {Calculating the exact \emph{Euclidean distance} from a point to an ellipsoid requires solving a sixth-order equation. We present a simple derivation on the exact computation in the supplemental material}. To circumvent this issue, Taubin~\cite{taubin1991estimation} uses the second-order Taylor expansion to approximate the orthogonal distance, \revise{whereas} Sampson~\cite{sampson1982fitting} weights the algebraic distance by the first-order differential.
\revise{However, geometric fitting usually requires proper initialization (from algebraic fitting)}, and it is also vulnerable to outliers because the objective function (Eq.~\ref{eq:gf}) is based on the LS principle. 
\revise{Later, \textit{iteratively re-weighted least-squares} (IRLS) is introduced to depress outliers, by which \textit{M-estimators} (robust kernels), such as Tukey~\cite{rousseeuw1991tutorial} and Huber~\cite{huber2004robust}, are used to reduce the effect of large residuals. IRLS is more stable and robust than ordinary least-squares in an outlier-contaminated environment. 
}

\section{Methodology} \label{sec:method}

\noindent{\textbf{Analysis of the input data.}}
For the given data points $\mathbf{X}=\{\mathbf{x}_i\in \mathbb{R}^3\}_{i=1}^{N}$, suppose $\mathbf{X}\sim p(\mathbf{x})$, \ie, $\mathbf{X}$ satisfies the probability distribution $p(\mathbf{x})$, then we use the KDE to model the point density by $p(\mathbf{x})=\frac{1}{N}\sum_{i=1}^{N}K_h(\mathbf{x}-\mathbf{x}_i),$ where $K_h(\mathbf{x}-\mathbf{x}_i)$ is the \emph{kernel function}, and $h$ is the \emph{kernel bandwidth}. A universal kernel is Gaussian function, which gives rise \revise{to} the following Gaussian mixture model: 
\begin{eqnarray}
	p(\mathbf{x})=\frac{1}{N}\sum_{i=1}^{N}\frac{1}{(2\pi h^2)^{d/2}}\exp(-\frac{\|\mathbf{x}-\mathbf{x}_i\|^2}{2h^2}),
	\label{eq:GMM}
\end{eqnarray}
where $d$ denotes the dimension ($d=3$ in our case). \revise{Despite} that Gaussian kernel function is broadly used, the choice of a globally suitable $h$ is not easy~\cite{bishop2006pattern}. To ease this problem, from the theory in~\cite{tang2017local} we adopt the local region for density estimation.

The $k$-nearest neighbour of $\mathbf{x}_i, i=1, \cdots, N$, is denoted as {\footnotesize{$\textrm{N}(\mathbf{x}_i)=\{\textrm{N}_1(\mathbf{x}_i), \textrm{N}_2(\mathbf{x}_i), \cdots, \textrm{N}_k(\mathbf{x}_i)\}.$}} Then the density at $\mathbf{x}_i$ is calculated by $p(\mathbf{x}_i)=\frac{1}{k+1}\sum_{\mathbf{x}\in \textrm{N}(\mathbf{x}_i)\cup \{\mathbf{x}_i\}}\frac{1}{(2\pi h^2)^{d/2}}\exp(-\frac{\|\mathbf{x}-\mathbf{x}_i\|^2}{2h^2}).$ We leverage \emph{kd-tree}~\cite{bentley1975multidimensional} to reduce the computational complexity from $O(N^2)$ to $O(N\log N)$. Different from~\cite{tang2017local} utilizing the same local $h$, we associate location $\mathbf{x}_i$ in the data space with kernel bandwidth $h$ by adaptively calculating the local covariance $h=\frac{1}{k}\sum_{\mathbf{x}\in\textrm{N}(\mathbf{x}_i)}(\mathbf{x}-\mathbf{x}_i)^T(\mathbf{x}-\mathbf{x}_i).$

After the density estimation of each point $\mathbf{x}_i\in \mathbf{X}$, we adopt the \emph{relative density-based outlier score} (RDOS)~\cite{tang2017local} to measure the extent of point $\mathbf{x}_i$, differing from its neighbourhood $\textrm{N}(\mathbf{x}_i)$, according to the following ratio
\begin{eqnarray}
	\RDOS(\mathbf{x}_i)=\frac{\sum_{\mathbf{x}\in \textrm{N}(\mathbf{x}_i)}p(\mathbf{x})}{|\textrm{N}(\mathbf{x}_i)|p(\mathbf{x}_i)}.
\end{eqnarray}
Intuitively, a larger $\RDOS(\mathbf{x}_i)$ indicates that $\mathbf{x}_i$ is outside a dense region. Thus it is more likely to be an outlier; otherwise, $\mathbf{x}_i$ can be deemed as non-outlier. We further use Lemma~\ref{lemma_1} to attain a quantitative analysis.
\begin{lemma}\label{lemma_1}
	Let the points $\mathbf{X}=\{\mathbf{x}_i\in \mathbb{R}^3\}_{i=1}^N$ be sampled from a continuous density distribution and the kernel function $K_h(x)$ be non-negative everywhere and integrated to one. Then, $\RDOS(\mathbf{x}_i)$ equals 1 with probability 1:
	\begin{eqnarray}
		\lim_{N \longrightarrow \infty}P(\RDOS(\mathbf{x}_i)=1)=1.
	\end{eqnarray}
\end{lemma}
Lemma \ref{lemma_1} provides a lower bound for outlier recognition. When $0<\RDOS(\mathbf{x}_i)<1$ or $\RDOS(\mathbf{x}_i)\approx 1$, we say that $\mathbf{x}_i$ is not an outlier, and $\mathbf{x}_i$ is possibly an outlier only if $\RDOS(\mathbf{x}_i)\gg 1$. \revise{The adaptive $\RDOS(\mathbf{x}_i)$ is introduced for the weight initialization of our method. Meanwhile it can also be used for ellipsoid modeling, as presented in the following.}

\noindent{\textbf{Ellipsoid modeling.}}
Suppose the given point set $\mathbf{X}=\{\mathbf{x}_i\in \mathbb{R}^3\}_{i=1}^{N}$ is fitted by an ellipsoid $e$ with the shape parameter $\bm{\theta}=(x_0, y_0, z_0, a, b, c, \alpha, \beta, \gamma)$, where $(x_0, y_0, z_0)$ is the ellipsoid center, $(a, b, c)$ are the three semi-axis \revise{lengths}, and $(\alpha, \beta, \gamma)$ are the Euler angles along the $x, y, \text{and}~ z$ axes. 
To attain the ellipsoid, we first create a unit sphere $s$ containing points $\mathbf{Y}=\{\mathbf{y}_m\in \mathbb{R}^3\}_{m=1}^{M}$ defined as
\begin{eqnarray}
x_m=x_c+\cos\theta_i\cdot\sin\psi_j,\quad y_m=y_c+\cos\theta_i\cdot\cos\psi_j,\quad z_m=z_c+\sin\theta_i, 
\end{eqnarray}
where $\mathbf{c}_s=(x_c, y_c, z_c)$ is the spherical center, $\theta\in[0, \pi)$, $\psi\in[0, 2\pi)$. To generate spherical points $\mathbf{y}_m$, \revise{the number of inliers can be counted as {\small{$M=\sum_{\mathbf{x}_i\in \mathbf{X}}\mathbbm{1}(\RDOS(\mathbf{x}_i)\leq 1)$}} (Lemma \ref{lemma_1})}, where $\mathbbm{1}$ is the indicator function. We relax the inlier constraint as \revise{$M=\sum_{\mathbf{x}_i\in\mathbf{X}}\mathbbm{1}(\RDOS(\mathbf{x}_i)\leq 2)$}, then  $\theta_i=\frac{\pi i}{[\sqrt{M}]}$ and $\psi_j=\frac{2\pi j}{[\sqrt{M}]}, i, j=1, \cdots, [\sqrt{M}]$, \revise{where $[\cdot]$ is a rounding function.}

Then a linear transformation $\mathcal{T}$ transforms the sphere $s$ to the real ellipsoid $e$ by $e=$$\mathcal{T}(s)=\mathbf{A}s+\mathbf{t},$ where $\mathbf{A}$ is the affine transformation matrix, and $\mathbf{t}$ is the translation vector. To solve $\mathbf{A}$ and $\mathbf{t}$, we formulate ellipsoid fitting as a likelihood estimation by first expressing the sphere model as a GMM with $M$ components,  $p(\mathbf{z})=\sum_{m=1}^{M}P(\mathbf{y}_m)p(\mathbf{z}|\mathbf{y}_m), \mathbf{z}\in \mathbb{R}^3,$ where $p(\mathbf{z}|\mathbf{y}_m)=\frac{1}{(2\pi \sigma^2)^{d/2}}\exp(-\frac{\|\mathbf{z}-\mathbf{y}_m\|^2}{2\sigma^2})$ is the Gaussian distribution and $P(\mathbf{y}_m)$ represents the probability selecting the component $\mathbf{y}_m$. \revise{To depress outliers, we add an additional uniform distribution $p(\mathbf{z}|\mathbf{y}_{M+1})=\frac{1}{V}$ relative to the volume $V$ of the bounding box of $\mathbf{X}$: $p(\mathbf{z})=w\frac{1}{V}+(1-w)\sum_{m=1}^{M}P(\mathbf{y}_m)p(\mathbf{z}|\mathbf{y}_m)$, where $w\in[0, 1]$ is the weight to balance the two distributions.}

Given that the spherical points $\mathbf{Y}=\{\mathbf{y}_m\in \mathbb{R}^3\}_{m=1}^{M}$ are generated uniformly, we set equal membership probability $P(\mathbf{y}_m)=\frac{1}{M}$ and isotropic covariance $\sigma^2$ for all components
{\begin{small}
	\begin{eqnarray}
		\begin{aligned}
			p(\mathbf{z})\!=\!\frac{w}{V}\!+\!\frac{1\!-\!w}{M}\sum_{m=1}^{M}(2\pi\sigma^2)^{-d/2}\exp(-\frac{1}{2\sigma^2}\|\mathbf{z}\!-\!\mathbf{y}_m\|^2).
		\end{aligned}
	\end{eqnarray}
\end{small}
}

The likelihood function $F(\Omega)=\prod_{i=1}^{N}p(\mathbf{x}_i|\Omega)$ of the input data  $\mathbf{X}=\{\mathbf{x}_i\in \mathbb{R}^3\}_{i=1}^N$ is maximized based on the independent and identical distribution assumption, where {\footnotesize{$\Omega=\{\mathbf{A}, \mathbf{t}, \sigma^2, w\}$}} is the parameter set. In~\cite{myronenko2010point}, the weight $w$ is preset as a hyper-parameter and tuned by users. However, we \revise{make no assumptions on the noise or outlier magnitude}. We take $w$ as a variable and automatically update it to find the optimal value. Finally, maximizing $F(\Omega)$ is equivalent to minimizing the following negative log-posterior
\revise{\begin{small}
	\begin{eqnarray}
		\begin{aligned}
			&E(\Omega|\mathbf{X})=-\sum_{i=1}^{N}\log p(\mathbf{x}_i|\Omega)
			=-\sum_{i=1}^{N}\log (\sum_{m=1}^{M}\frac{1-w}{M}\frac{1}{(2\pi\sigma^2)^{d/2}}\exp({\frac{\|\mathbf{x}_i-(\mathbf{A}\mathbf{y}_m+\mathbf{t})\|^2}{2\sigma^2}})+\frac{w}{V}).
		\end{aligned}
	\end{eqnarray}
\end{small}
}
\section{EM Algorithm}
We adopt the EM framework~\cite{moon1996expectation} for ellipsoid fitting. The basic idea behind is first guessing an "old" parameter $\Omega^{old}$ and then use the Bayesian theorem~\cite{joyce2003bayes} to compute a posterior probability or responsibility of the mixture components, which is the expectation or E-step of the algorithm. In the subsequent maximization or M-step, the "new" parameter $\Omega$ is updated by minimizing the expectation of the completed-data negative log-likelihood $Q$ function (detailed in the supplemental material). The update of EM is detailed as follows.\\

\noindent{\textbf{E-step:}} We compute the posterior probability regarding the uniform distribution and each mixture component in GMM, respectively. 
\makeatletter\def\@captype{figure}\makeatother
\begin{footnotesize}
	\begin{eqnarray}
		\begin{aligned}
			&p^{old}(\revise{\mathbf{y}_{M+1}}|\mathbf{x}_i, \Omega)=
			\frac{\frac{w}{V}}{w\frac{1}{V}+(1-w)\sum_{k=1}^{M}\frac{1}{M}p(\mathbf{x}_i|\mathbf{y}_k)}
			&p^{old}(\mathbf{y}_m|\mathbf{x}_i, \Omega)=
			\frac{\frac{1-w}{M}\frac{1}{(2\pi\sigma^2)^{d/2}}\exp({\frac{-\|\mathbf{x}_i- \mathbf{A}^{old}\mathbf{y}_m+\mathbf{t}^{old})\|^2}{2\sigma^2}})}{w\frac{1}{V}+(1-w)\sum_{k=1}^{M}\frac{1}{M}p(\mathbf{x}_i|\mathbf{y}_k)}\\
			&=\frac{1}{1+\frac{V}{M}\frac{1-w}{w}\sum_{k=1}^{m}\exp({\frac{-\|\mathbf{x}_i-(\mathbf{A}^{old}\mathbf{y}_k+\mathbf{t}^{old})\|^2}{2\sigma^2}})},
			&=\frac{\exp({\frac{-\|\mathbf{x}_i-( \mathbf{A}^{old}\mathbf{y}_k+\mathbf{t}^{old})\|^2}{2\sigma^2}})}{\sum_{k=1}^{m}\exp({\frac{-\|\mathbf{x}_i-( \mathbf{A}^{old}\mathbf{y}_k+\mathbf{t}^{old})\|^2}{2\sigma^2}})+(2\pi\sigma^2)^{d/2}\frac{w}{1-w}\frac{M}{V}}.
		\end{aligned}
	\end{eqnarray}
\end{footnotesize}

\noindent{\textbf{M-step:}} We update all parameters in $\Omega$ by minimizing $Q(\Omega, \Omega^{old})$. We take partial derivatives of $Q$ with respect to each parameter and equate them to zero. Solving $\frac{\partial Q}{\partial \mathbf{t}}=0$, we attain $\mathbf{t}=\frac{1}{N_p}(\mathbf{X}^{T}\mathbf{P}^T\mathbf{1}-\mathbf{A}\mathbf{Y}^{T}\mathbf{P}\mathbf{1})$, where $\mathbf{X}=[\mathbf{x}_1, \cdots, \mathbf{x}_N]^T$ and $\mathbf{Y}=[\mathbf{y}_1, \cdots, \mathbf{y}_M]^T$. $\mathbf{P}$ is the correspondence probability matrix with elements $p_{mn}=p^{old}(\mathbf{y}_m|\mathbf{x}_n)$, and $\mathbf{1}$ is the unit column vector. Similarly, $w=\frac{N_o}{N_p+N_o},$ $\mathbf{A}=(\mathbf{\hat{X}}^T\mathbf{P}^T\hat{\mathbf{Y}})(\hat{\mathbf{Y}}^T\diag(\mathbf{P}\mathbf{1})\hat{\mathbf{Y}})^{-1},$ and $\sigma^2=\frac{1}{N_pd}\tr(\mathbf{\hat{X}^T}(\diag(\mathbf{P}^T\mathbf{1})\mathbf{\hat{X}}$ $-\mathbf{P}^T\hat{\mathbf{Y}}\mathbf{A}^T))$, where $\hat{\mathbf{X}}=\mathbf{X}-\frac{1}{N_p}\mathbf{O}\mathbf{P}\mathbf{X}$, $\hat{\mathbf{Y}}=\mathbf{Y}-\frac{1}{N_p}\mathbf{O}\mathbf{P}^T\mathbf{Y}$. $\mathbf{O}=\mathbf{1}\mathbf{1}^T$ is all ones matrix, and $\diag(\mathbf{a})$ is the diagonal matrix formed by vector $\mathbf{a}$.

Furthermore, we adopt an $\varepsilon$-accelerated technique~\cite{wang2008acceleration} in our method to speed up the EM convergence. To this end, we formalize the total parameters in $\Omega$ as a $1 \times 14$ vector denoted by $\boldmath{\Omega}$. Then, the update of the new sequence $\{\dot{\boldmath{\Omega}}^{(n)}\}_{n\geq 0}$ is 
\begin{eqnarray}
	\dot{\boldmath{\Omega}}^{(n)}=\boldmath{\Omega}^{(n+1)}\!+\!((\boldmath{\Omega}^{(n+2)}\!-\!\boldmath{\Omega}^{(n+1)})^{-1}\!-\!(\boldmath{\Omega}^{(n+1)}\!-\!\boldmath{\Omega}^{(n)})^{-1})^{-1},
\end{eqnarray}
where the inverse of a vector $\mathbf{x}$ is defined as $[\mathbf{x}]^{-1}=\mathbf{x}/\|\mathbf{x}\|^2$. The above steps are repeated until
\begin{eqnarray}
	{||\dot{\boldmath{\Omega}}^{(n+1)}-\dot{\boldmath{\Omega}}^{(n)}||}^{2}\leq\delta,
\end{eqnarray}
where $\delta=10^{-8}$ is the default convergence accuracy.\\

\noindent{\textbf{Ellipsoid parameter.}}~Once we attain the optimal affine matrix  $\hat{\mathbf{A}}$ \revise{(the rotation matrix and the scales can be recovered from it)} and the translation vector $\hat{\mathbf{t}}$ by the $\varepsilon$-accelerated EM algorithm, the spherical point $\mathbf{y}_{m}\in s, m=1, \cdots, M$, becomes $\mathbf{x}=\hat{\mathbf{A}}\mathbf{y}_{m}+\hat{\mathbf{t}}$, \revise{where $\mathbf{x}\in e$ on the ellipsoid is no longer homogeneous. However, we lay more emphasis on the 
nine geometric parameters} of an ellipsoid (derived in the supplemental material), which are expressed as

\begin{footnotesize}	
\begin{eqnarray}
	\centering
	\boxed{
		\begin{aligned}
			\hat{\mathbf{c}}_e=\hat{\mathbf{t}}+\Hat{\mathbf{A}}\mathbf{c}_{s},\ \hat{a}=\sqrt{\lambda_1},\ \hat{b}=\sqrt{\lambda_2},\ \hat{c}=\sqrt{\lambda_3},\
			\hat{\alpha}=\atan2{\frac{-\mathbf{Q}_{31}}{\sqrt{(\mathbf{Q}_{11}+\mathbf{Q}_{21})^2}}},\ \hat{\beta}=\atan2{\frac{\mathbf{Q}_{21}}{\mathbf{Q}_{11}}}, \ \hat{\gamma}=\atan2{\frac{\mathbf{Q}_{32}}{\mathbf{Q}_{33}}},\
		\end{aligned}
	}
\end{eqnarray}
\end{footnotesize}where $\lambda_1$, $\lambda_2$, $\lambda_3$, and $\mathbf{Q}_{3 \times 3}$ are the eigenvalues and the orthogonal matrix attained via eigen-decomposition of $\Hat{\mathbf{B}}=\hat{\mathbf{A}}\hat{\mathbf{A}}^T$.

\section{Experiments}\label{sec:experiment}
In this section, the performance of the proposed method is \revise{tested} and compared with seven representative approaches falling into three categories, \ie, algebraic methods: DLS~\cite{li2004least}, HES~\cite{kesaniemi2017direct}, MQF~\cite{birdal2019generic} and Koop~\cite{vajk2003identification}; geometric methods: GF~\cite{bektas2015least} and Taubin~\cite{taubin1991estimation}; and the robust one: RIX~\cite{lopez2017robust} dedicated for outlier handling. Furthermore, we demonstrate the applications of the proposed method for 3D scanned point clouds, where outliers, noise, and occlusion exist.
\revise{For numerical stability~\cite{hartley1997defense}, the input data $\mathbf{X}=\{\mathbf{x}_i\in \mathbb{R}^3\}_{i=1}^{N}$ is first normalized. We use $\mathbf{A}=\mathbf{I}, \mathbf{t}=\mathbf{0}, w=\frac{\#\{\mathbf{x}_i|\RDOS(\mathbf{x}_i)>2\}}{N}$ and $\sigma^2=\frac{1}{DNM}\sum_{i, m}^{N, M}\|\mathbf{x}_i-\mathbf{y}_m\|^2$ to initialize $\boldmath{\Omega}$ in the EM algorithm. The weight in MQF is 0.3, as suggested by the authors. The maximal step size of RIX is tuned from 50 to 100, whereas the minimal one is 0.001. The scale factor of RIX is tuned from 1.5 to 6 as fixed values often lead to noticeable deviations}. Similar to~\cite{kesaniemi2017direct, lopez2017robust}, the fitting accuracy is assessed through the offset error $E_{\mathbf{c}}$ and the shape error $E_{\mathbf{a}}$ 
\begin{eqnarray}	
	E_{\mathbf{c}}=||\mathbf{c}_t-\hat{\mathbf{c}}||_2,~ E_{\mathbf{a}}=\frac{s_{\max}(\mathbf{\hat{A}^{-1}}\mathbf{A}_t)}{s_{\min}(\mathbf{\hat{A}^{-1}}\mathbf{A}_t)}-1, %
\end{eqnarray}
where $\mathbf{c}_t$ and $\hat{\mathbf{c}}$, $\mathbf{A}_t$ and $\hat{\mathbf{A}}$ are the offsets and the affine matrices of the ground truth and the fitted ellipsoids, respectively and $s_{\max}$ and $s_{\min}$ represent the largest and the smallest singular values of the residual transformation $\hat{\mathbf{A}}^{-1}\mathbf{A}_t$, respectively. For each test, we perform 100 independent trials, and the average metric is reported.\\

\begin{figure*}[t]
	\includegraphics[width=1\textwidth]{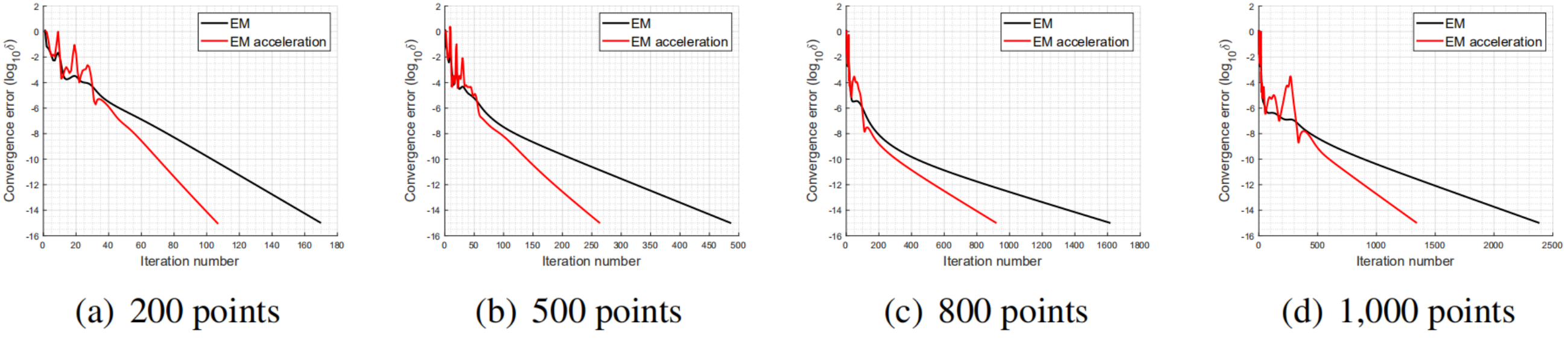}
	\vskip -0.3cm
	\caption{Convergence comparison between the original EM (black line) and the accelerated one (red line) under different point numbers. The acceleration effect becomes more significant as the number of points or the required fitting accuracy increase.
	}\label{fig:em_acc}
	\vskip -0.5cm
\end{figure*}

\vskip -0.3cm
\begin{table}[!htbp]
	\renewcommand\arraystretch{0.7}
	\centering
	\normalsize
	\caption{Comparisons of different methods on noisy data, where \textbf{bold font} is the top fitter.}
	\resizebox{0.95\textwidth}{1.4cm}{
		\begin{tabular}{c|ccccccccc}
			\toprule
			\diagbox{Noise (\%)}{Method}&Metric &DLS\cite{li2004least}&HES\cite{kesaniemi2017direct}&MQF\cite{birdal2019generic}&Koop\cite{vajk2003identification}&Taubin\cite{taubin1991estimation}&GF\cite{bektas2015least}&RIX\cite{lopez2017robust}&Ours\cr
			\cmidrule(lr){1-10}
			\multirow{2}{*}{5} &$E_\mathbf{c}$&3.43&3.42&3.47&4.06&3.89&1.31&\textbf{0.67}&1.03\\
			&$E_\mathbf{a}$&0.45&0.46&0.57&0.74&0.63&\textbf{0.14}&0.15&\textbf{0.14}\\ 
			\multirow{2}{*}{10}&$E_\mathbf{c}$ &3.92&3.90&4.14&5.20&4.83&1.90&\textbf{1.17}&1.33\\
			&$E_\mathbf{a}$&0.47&0.48&0.65&0.88&0.71&0.21&0.23&\textbf{0.20}\\
			\multirow{2}{*}{15} &$E_\mathbf{c}$&4.51&4.49&4.11&6.12&5.87&2.14&1.66&\textbf{1.58}\\
			&$E_\mathbf{a}$&0.48&0.49&0.76&1.11&0.86&0.29&0.29&\textbf{0.25}\\
			\multirow{2}{*}{20}&$E_\mathbf{c}$ &4.61&4.60&3.62&7.83&6.86&3.23&2.20&\textbf{2.01}\\
			&$E_\mathbf{a}$&0.46&0.47&0.84&1.43&0.94&0.53&0.33&\textbf{0.32}\\
			\multirow{2}{*}{25}&$E_\mathbf{c}$ &4.85&4.84&5.10&8.97&8.81&4.21&2.68&\textbf{2.16}\\
			&$E_\mathbf{a}$&0.44&0.46&1.09&1.65&1.18&0.94&\textbf{0.38}&0.40\\
			\bottomrule
		\end{tabular}
	}
	\label{tab:noise}
	\vskip -0.1cm
\end{table}
\noindent{\textbf{Effect of the $\varepsilon$ technique.}}~First, we reveal the effect of $\varepsilon$-accelerated EM for 200, 500, 800, and 1,000 data points. The results are reported in Fig.~\ref{fig:em_acc}, where some observations can be drawn: (1) for the fixed point number, the acceleration effect is more significant as the required convergence accuracy increases; (2) conversely, for fixed  accuracy, the acceleration effect is also more significant as the point number increases. Therefore, the $\varepsilon$ technique can effectively speed up the convergence of the ellipsoid fitting process, especially for points with a large magnitude under a high accuracy fitting requirement.\\

\begin{figure}
\centering
\includegraphics[width=0.63\textwidth]{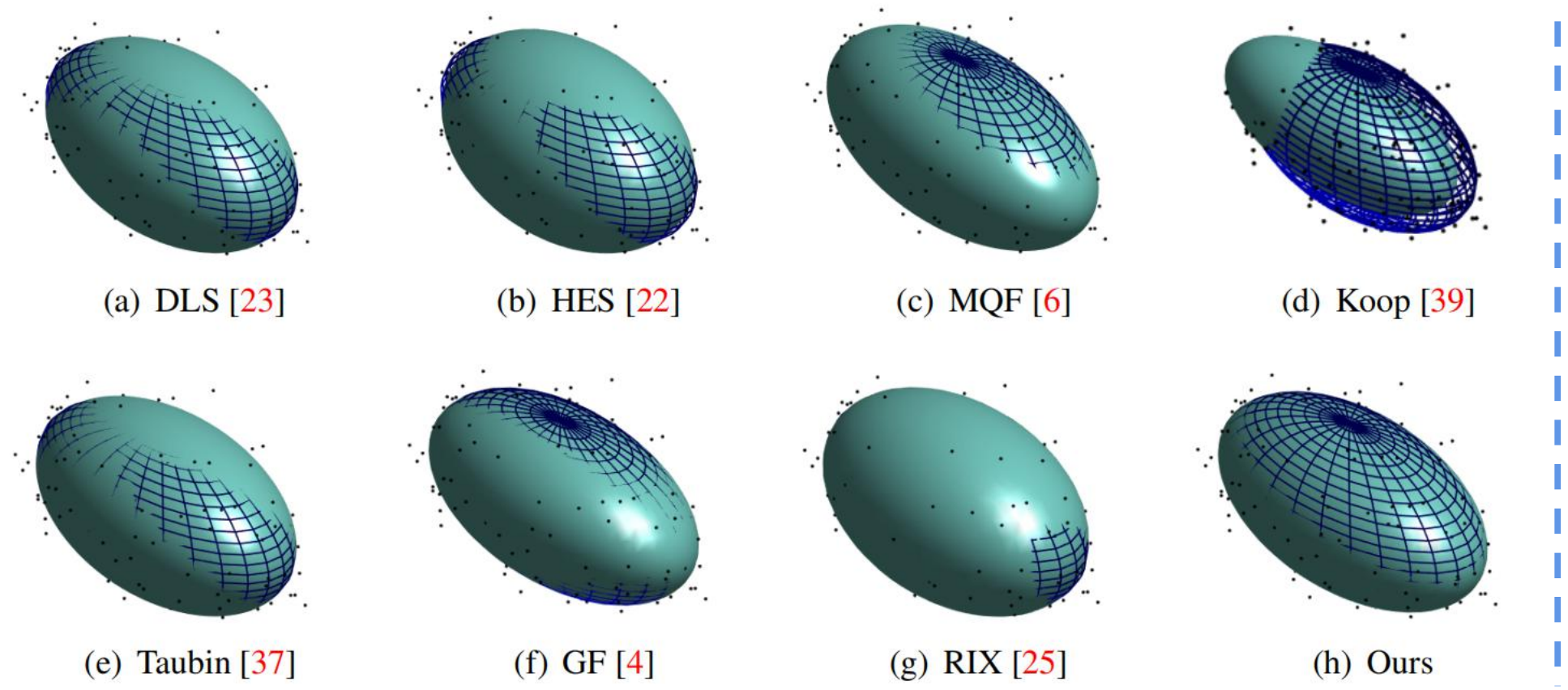}
\includegraphics[width=0.35\textwidth]{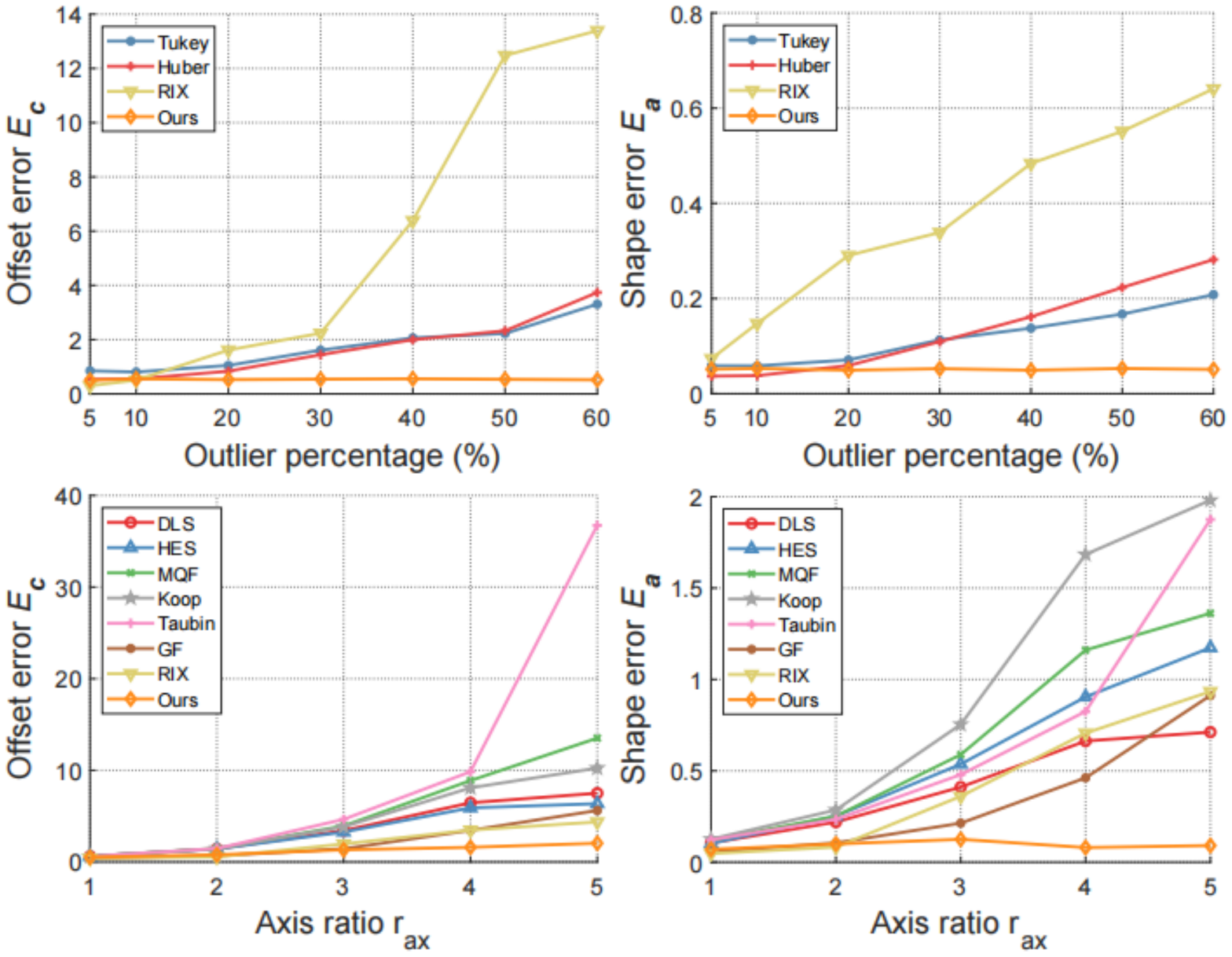}
\vskip -0.2cm
\caption{Left: Fitting results under 10\% Gaussian noise and 1\% outliers (net ellipsoid is the ground truth); right: Our method exhibits higher robustness against outliers and axis ratio.}\label{fig:noise_exam}
\vskip -0.3cm	
\vskip -0.3cm
\label{ellEx}
\end{figure}

\noin{Effect of noise.}~Next, we add different Gaussian noise with zero mean and \revise{standard deviation $\sigma\in[5\%, 25\%]$} to 200 data points. 
The average offset and shape deviations $E_\mathbf{c}$ and $E_\mathbf{a}$ are reported in Table~\ref{tab:noise}. As observed, our fit attains the overall best performance and is more robust when heavier noise is added. GF has minor deviations than the other LS-based methods. However, when noise goes up, see $\sigma>20\%$, a significant error exists, indicating its instability for severe noise. Koop attains the largest deviations among all methods, while DLS and HES share quite similar performance. As a robust method, RIX achieves the second-best performance, but with noise increasing, such as $\sigma>15\%$, it results in more offset errors than ours. Ellipsoid fitting examples are presented in the left panel of Fig.~\ref{ellEx}.\\

\begin{figure*}[h]
\vskip -0.8cm
\centering
\subfigure[Input]{
		\begin{minipage}[b]{0.174\textwidth}
			\includegraphics[width=1\textwidth]{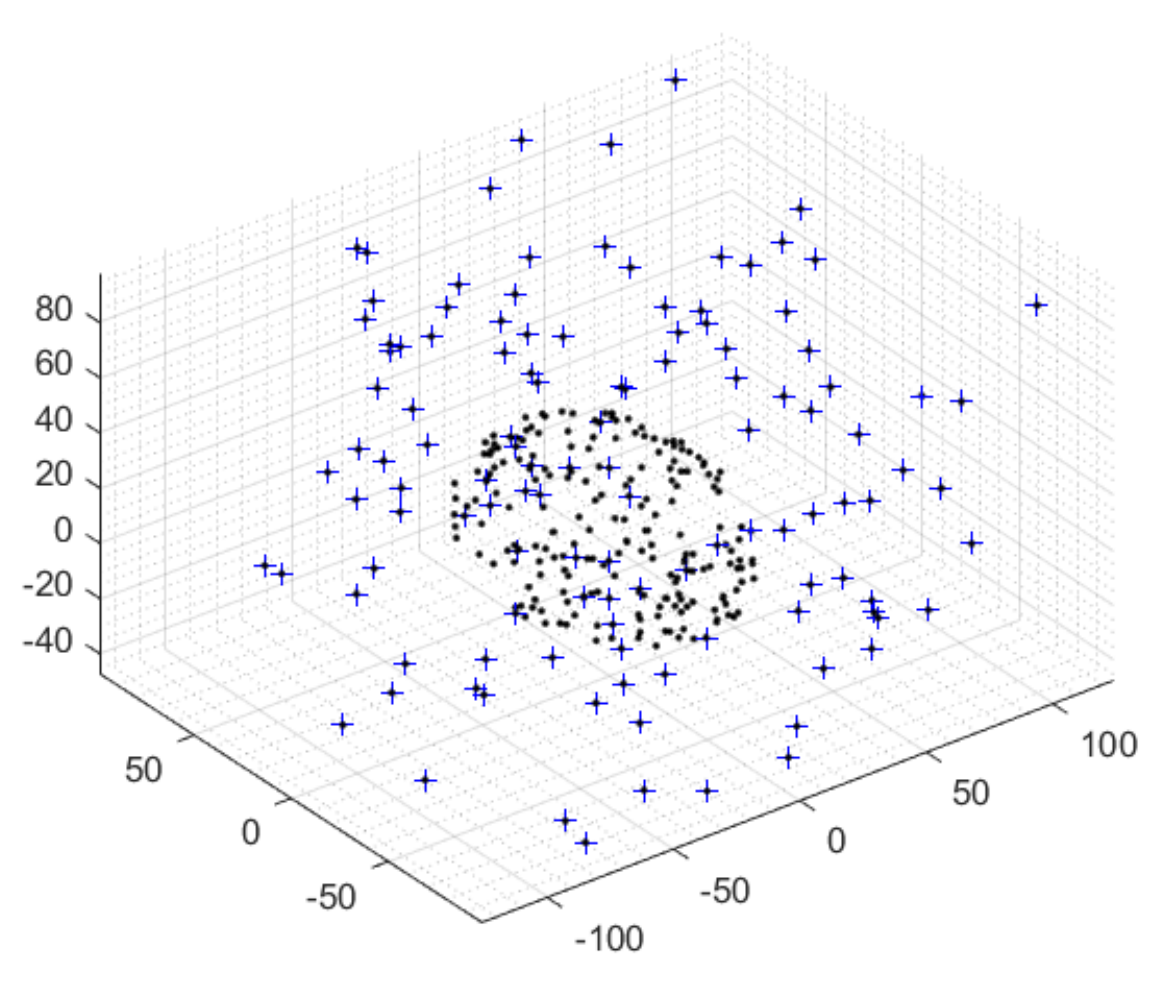}\\
			\includegraphics[width=1\textwidth]{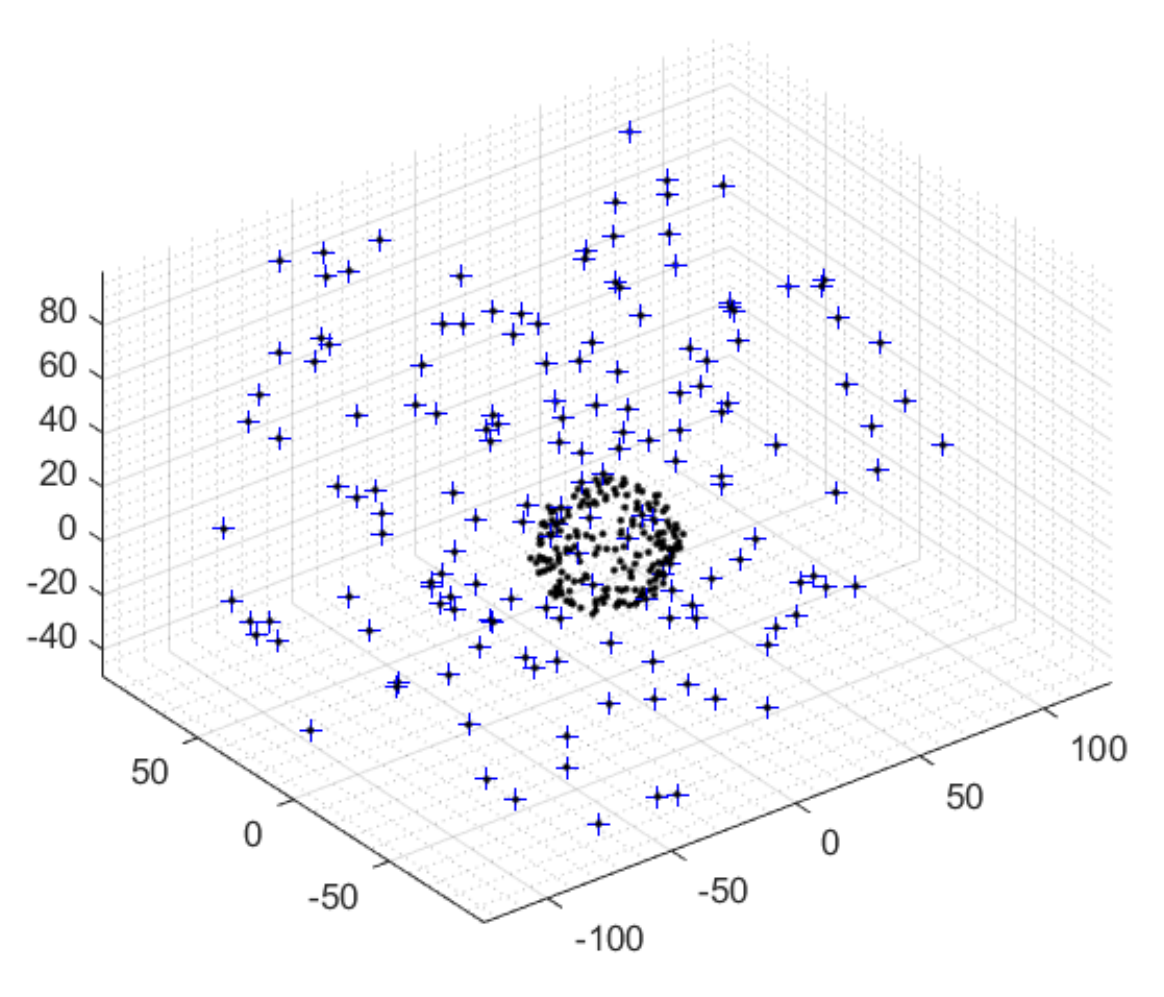}
		\end{minipage}
}
\subfigure[\revise{Tukey~\cite{rousseeuw1991tutorial}}]{
	\begin{minipage}[b]{0.174\textwidth}
		\includegraphics[width=1\textwidth]{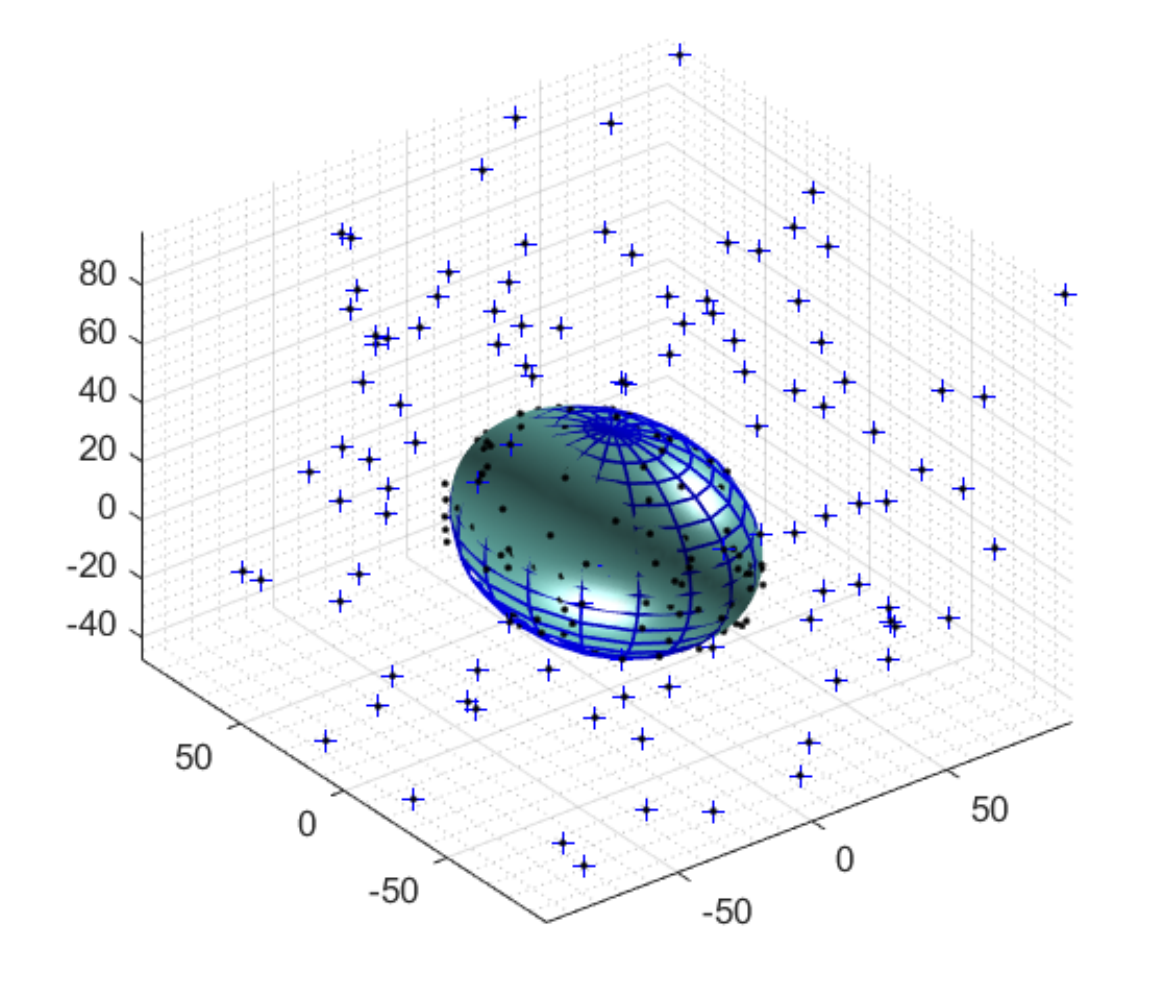}\\
		\includegraphics[width=1\textwidth]{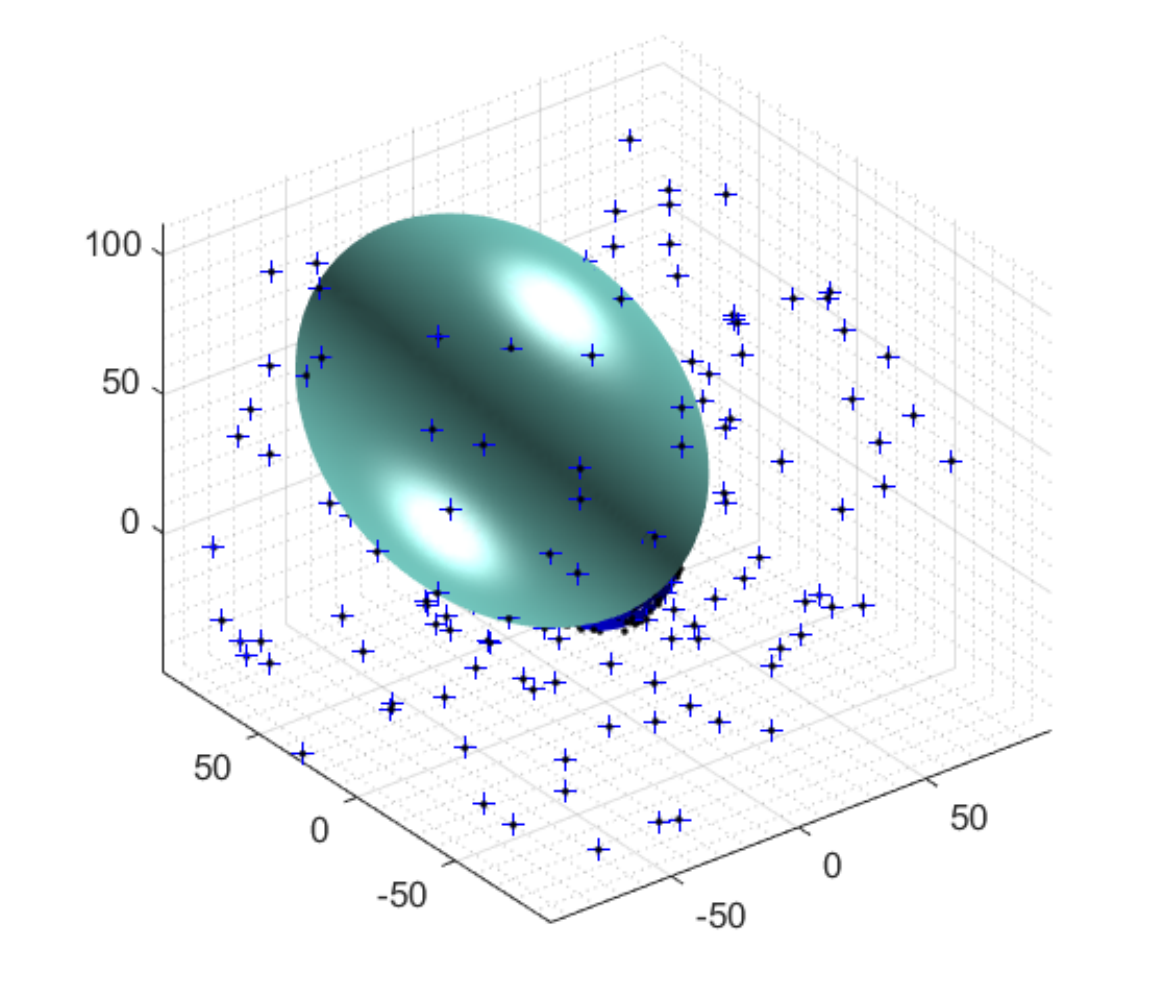}
	\end{minipage}
}
\subfigure[\revise{Huber~\cite{huber2004robust}}]{
	\begin{minipage}[b]{0.174\textwidth}
		\includegraphics[width=1\textwidth]{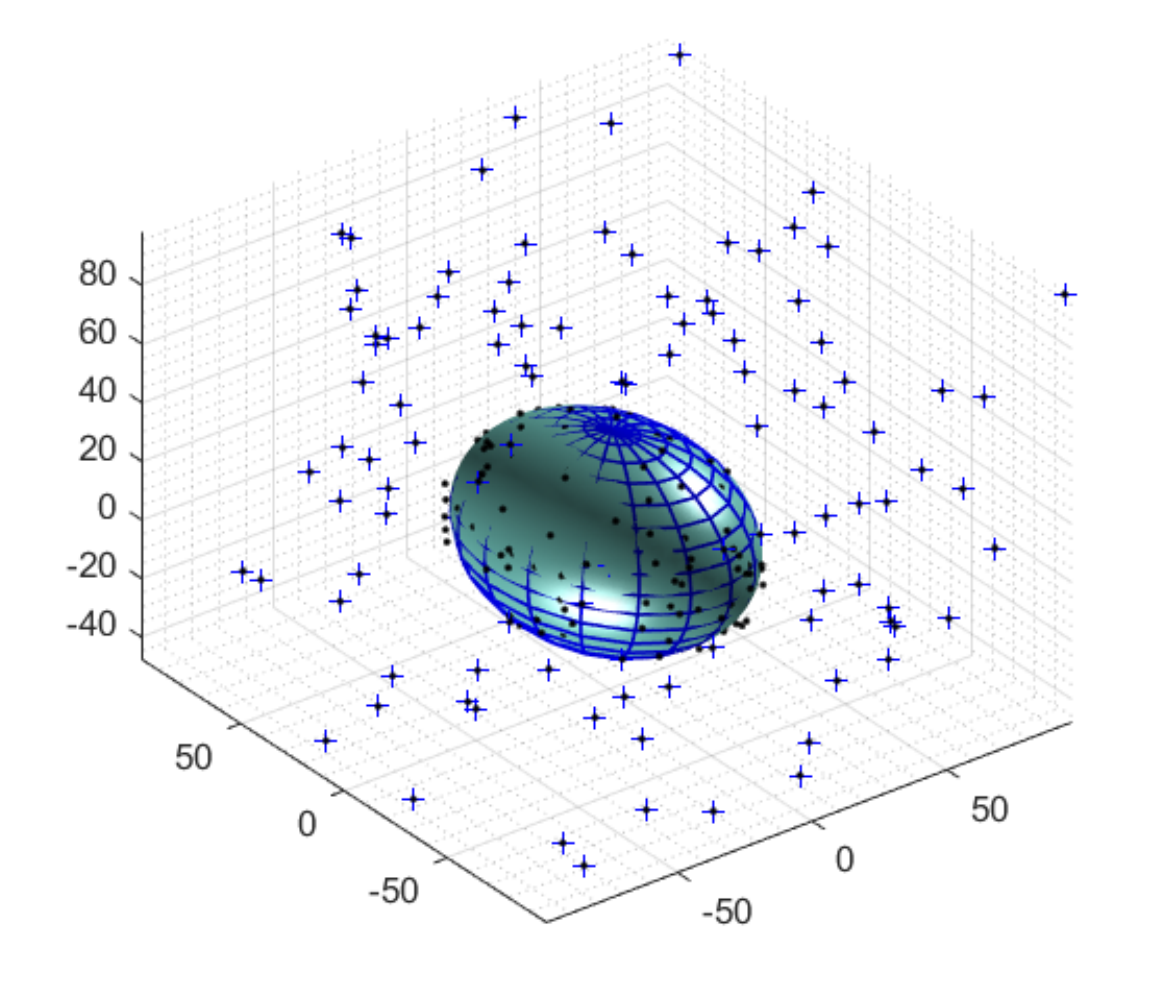}\\
		\includegraphics[width=1\textwidth]{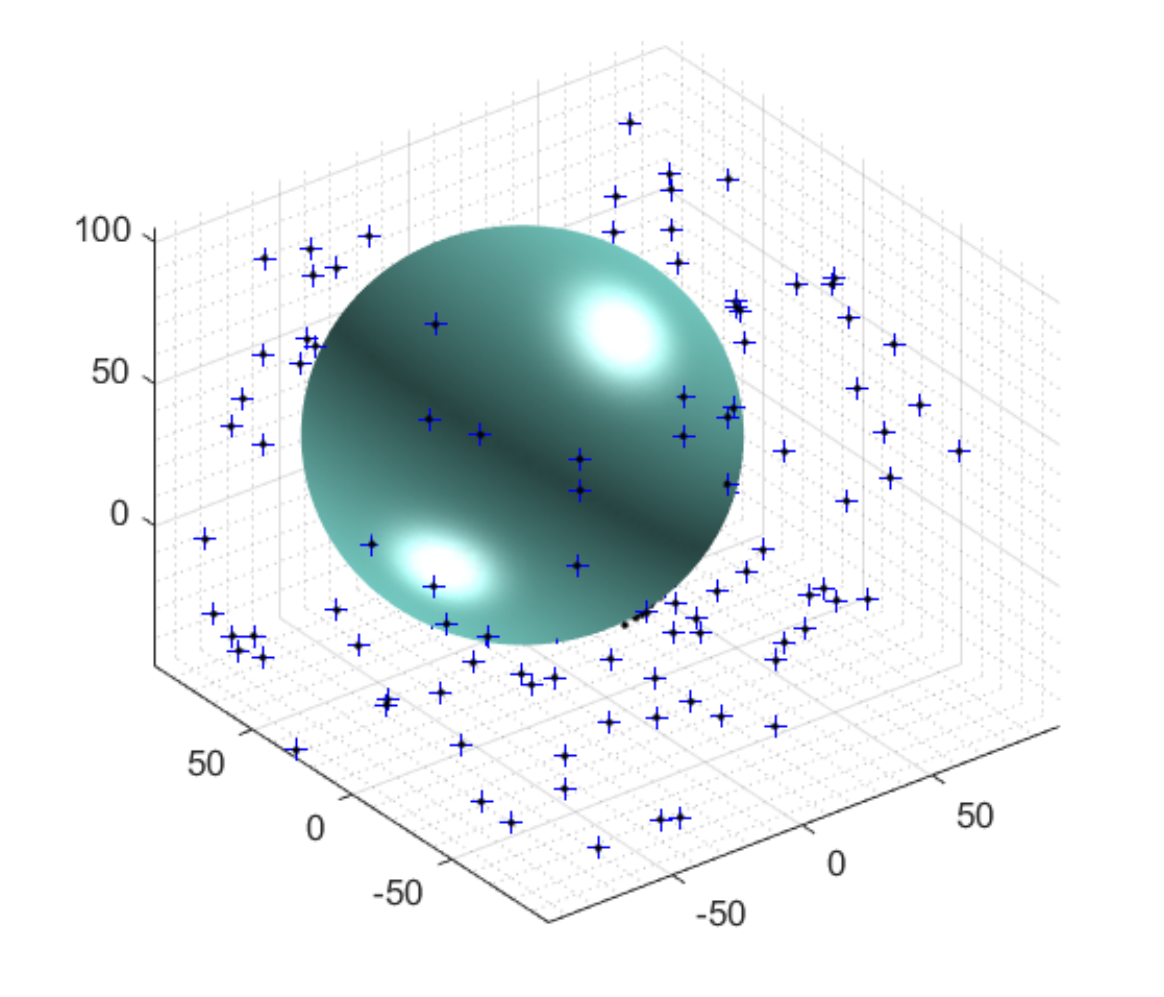}
	\end{minipage}
}
\subfigure[RIX~\cite{lopez2017robust}]{
	\begin{minipage}[b]{0.174\textwidth}
		\includegraphics[width=1\textwidth]{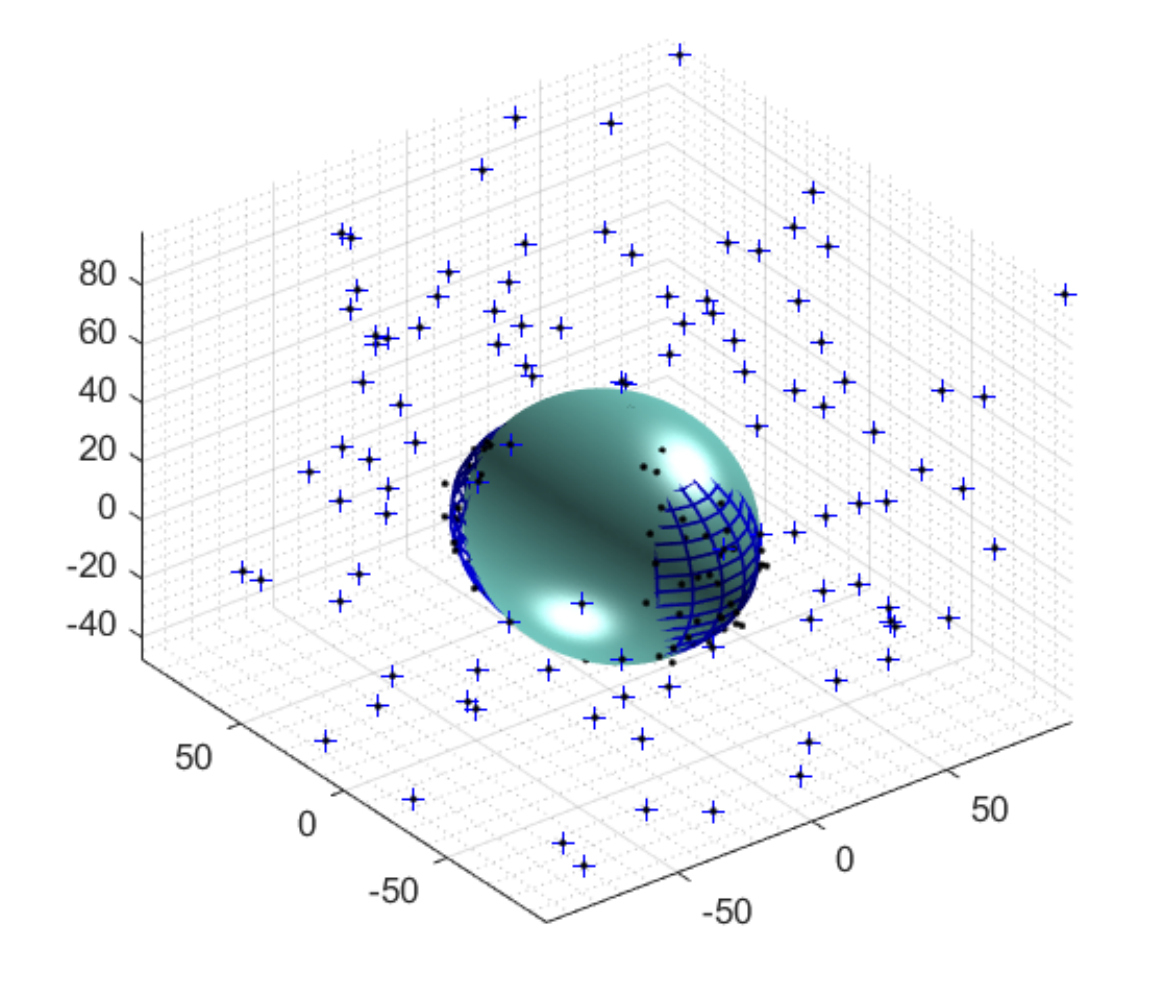}\\
		\includegraphics[width=1\textwidth]{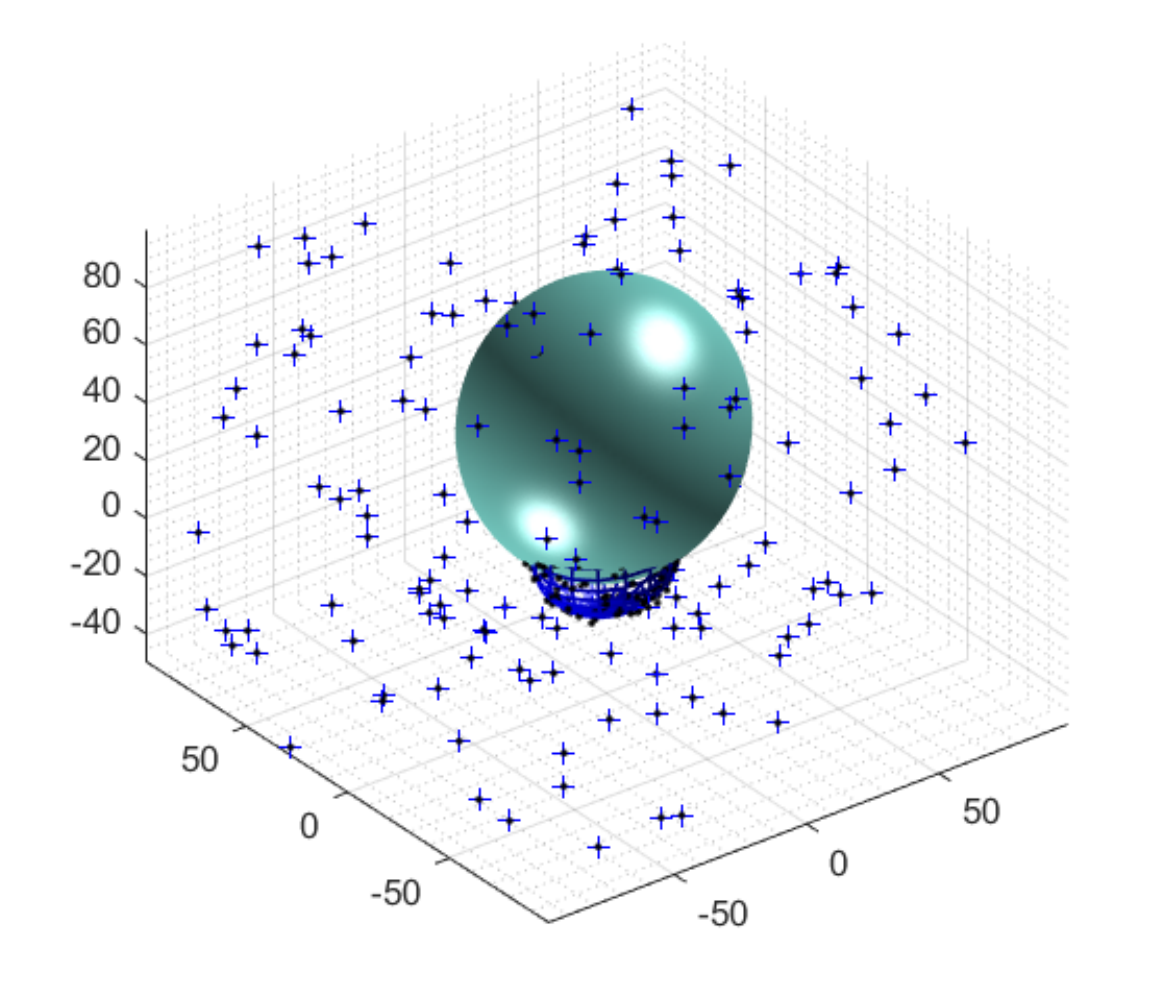}
	\end{minipage}
}
\subfigure[Ours]{
	\begin{minipage}[b]{0.174\textwidth}
		\includegraphics[width=1\textwidth]{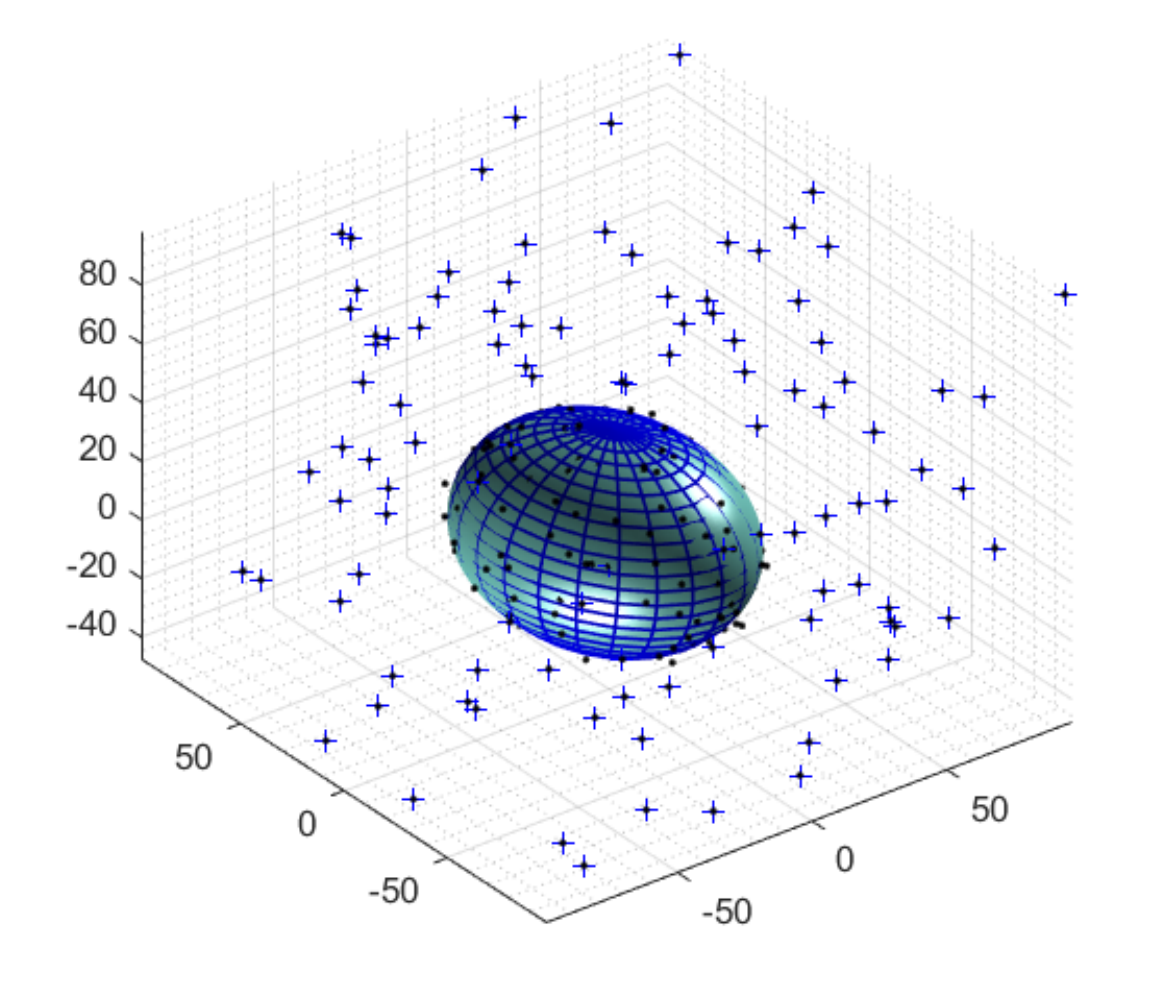}\\
		\includegraphics[width=1\textwidth]{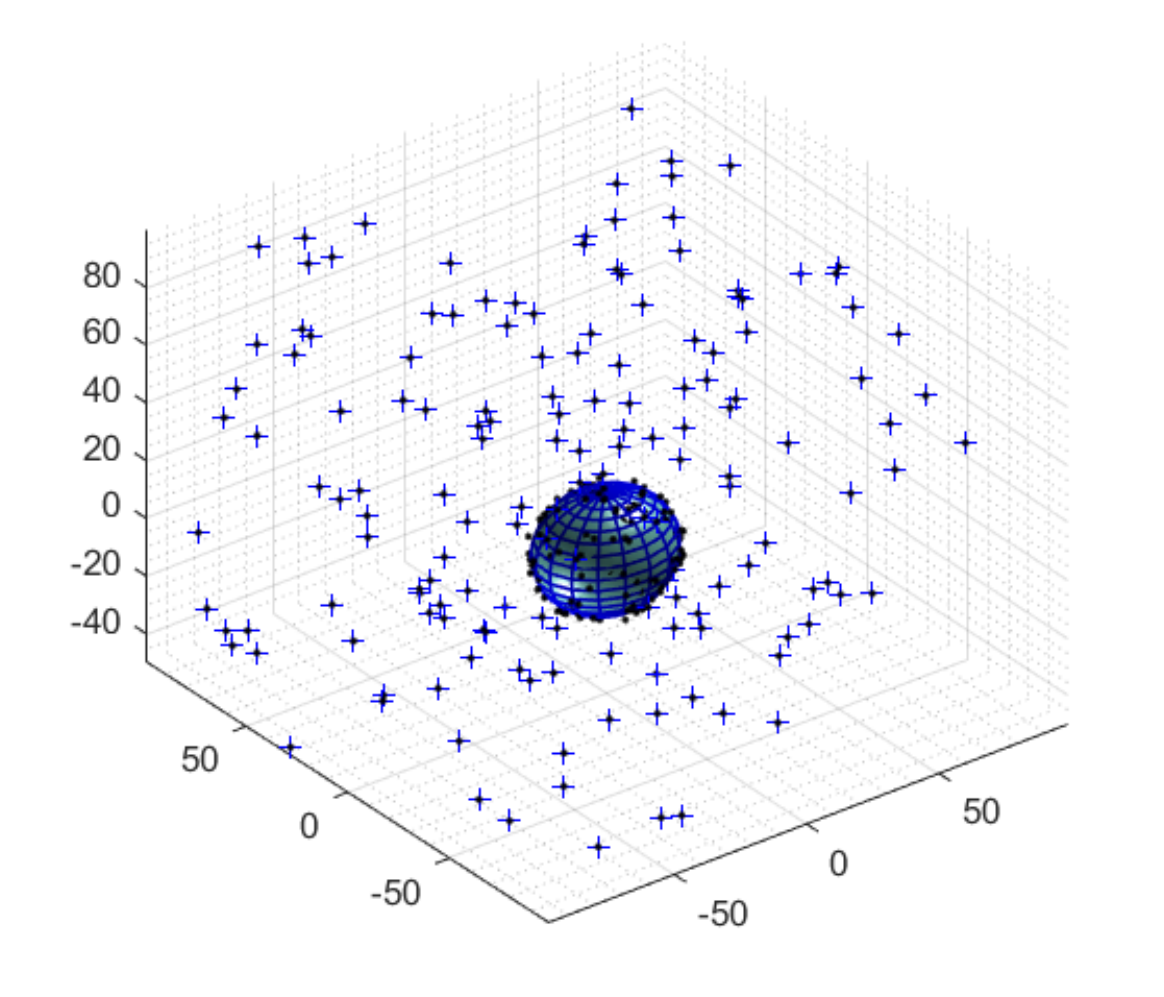}
	\end{minipage}
}
\vskip -0.1cm
\caption{\revise{Ellipsoid fitting examples in the presence of outliers ('\revise{+}'). As outliers increase from 60\% (first row) to 80\% (second row), the proposed method outperforms M-estimators of Tukey and Huber, and RIX with higher robustness.}}\label{fig:outlier_exam}
\label{fig:axis}
\vskip -0.3cm
\end{figure*}

\begin{figure*}[t]
	\vskip -0.25cm
	\centering
	\subfigure[\tiny{DLS~\cite{li2004least}}]{
		\begin{minipage}[b]{0.1\textwidth}
			\includegraphics[width=1\textwidth]{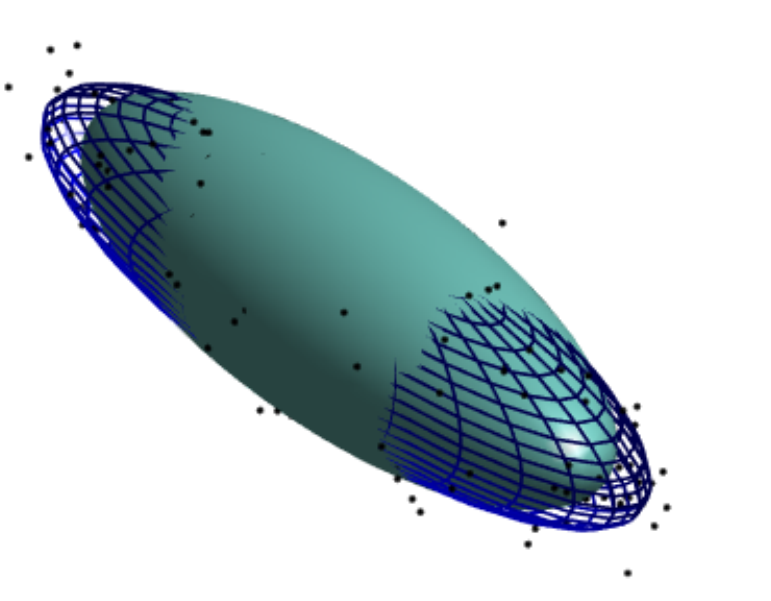}\\
			\includegraphics[width=1\textwidth]{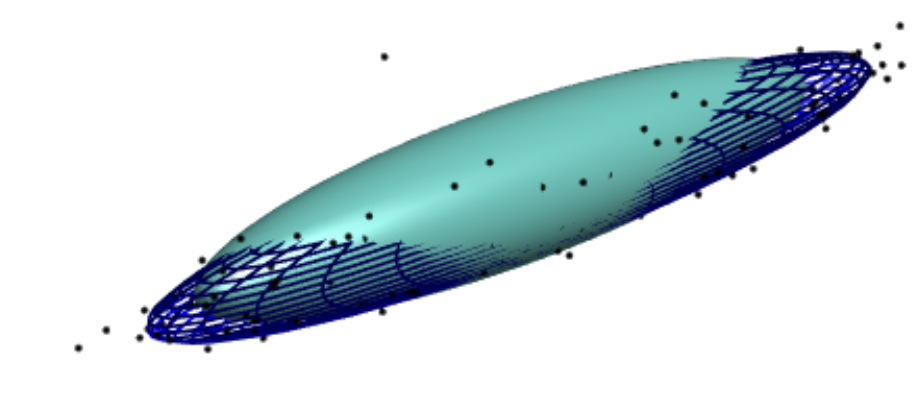}\\
			\includegraphics[width=1\textwidth]{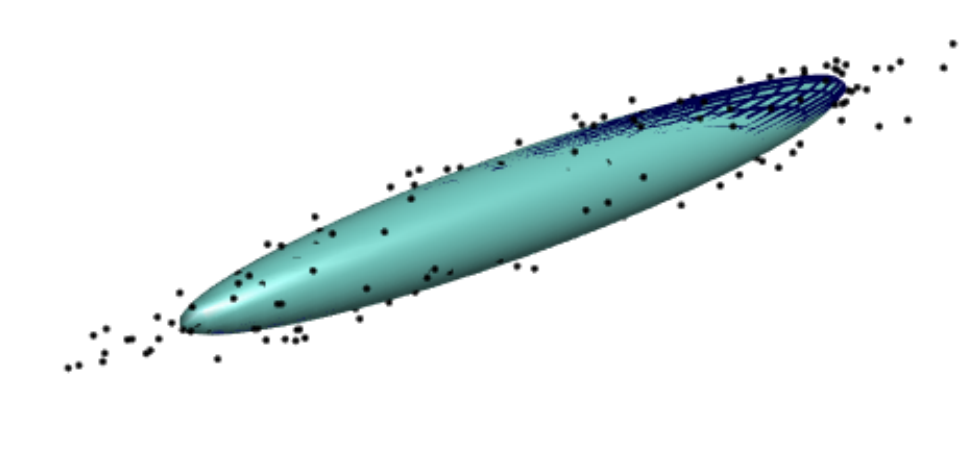}
		\end{minipage}
	}
	\subfigure[\tiny{HES~\cite{kesaniemi2017direct}}]{
		\begin{minipage}[b]{0.1\textwidth}
			\includegraphics[width=1\textwidth]{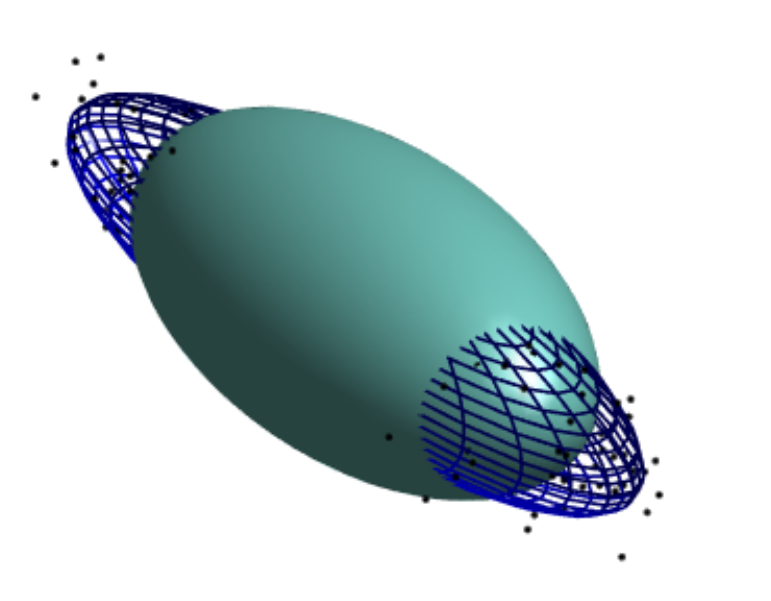}\\
			\includegraphics[width=1\textwidth]{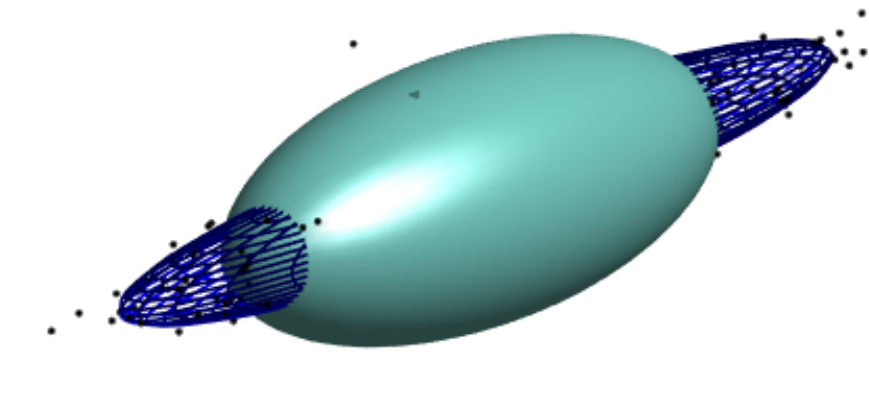}\\
			\includegraphics[width=1\textwidth]{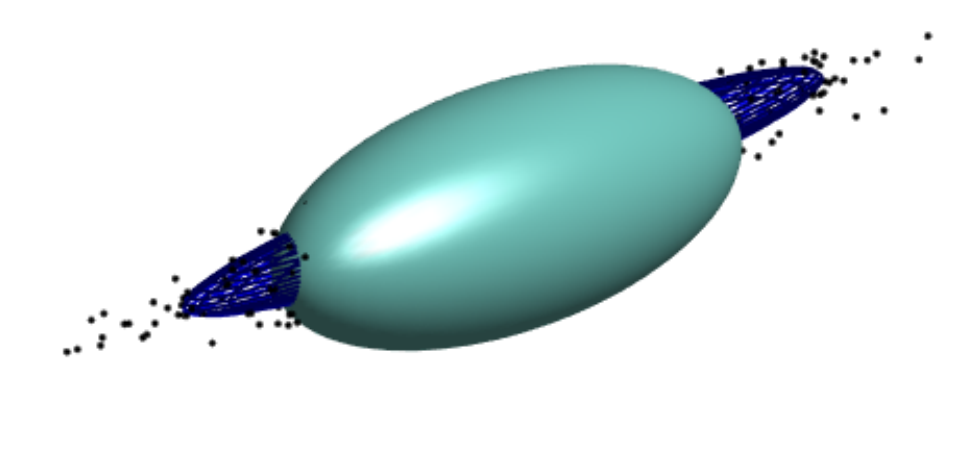}
		\end{minipage}
	}
	\subfigure[\tiny{MQF~\cite{birdal2019generic}}]{
		\begin{minipage}[b]{0.1\textwidth}
			\includegraphics[width=1\textwidth]{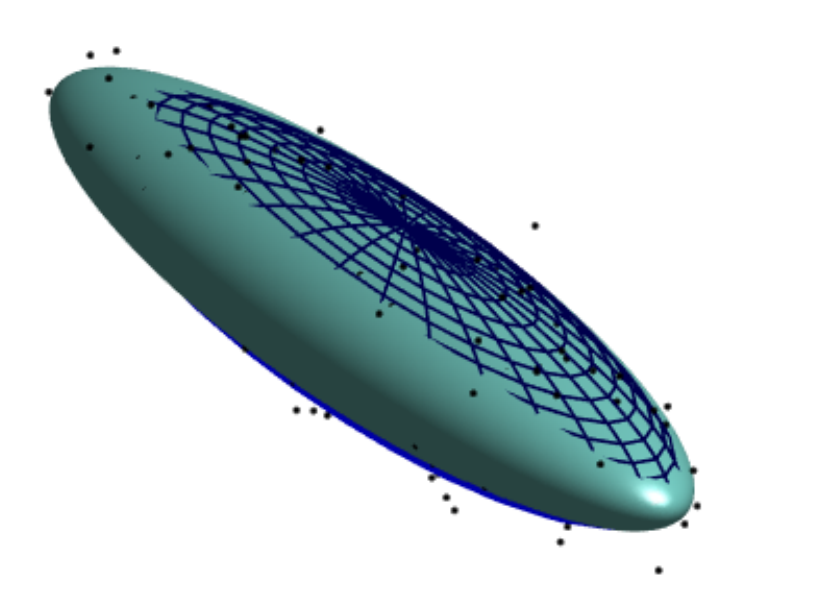}\\
			\includegraphics[width=1\textwidth]{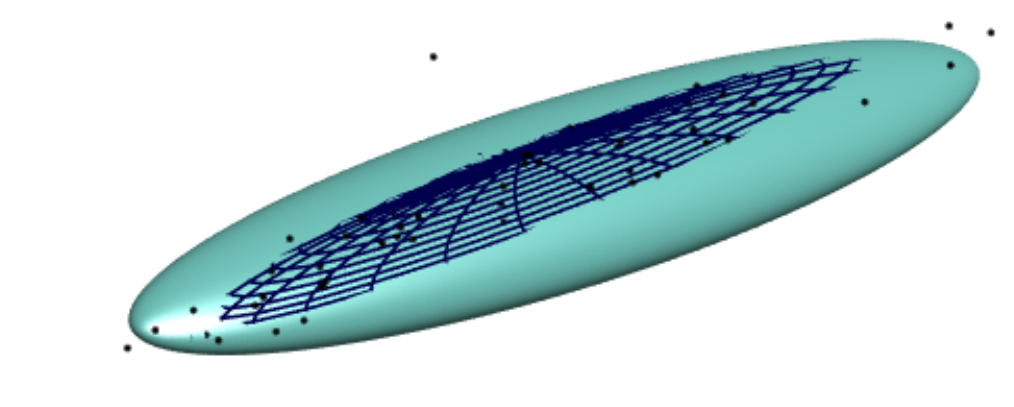}\\
			\includegraphics[width=1\textwidth]{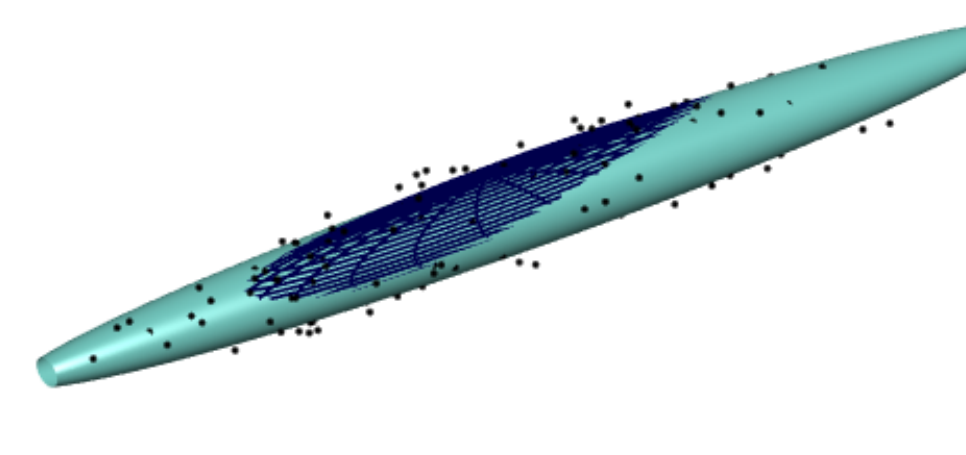}
		\end{minipage}
	}
	\subfigure[\tiny{Koop~\cite{vajk2003identification}}]{
		\begin{minipage}[b]{0.1\textwidth}
			\includegraphics[width=1\textwidth]{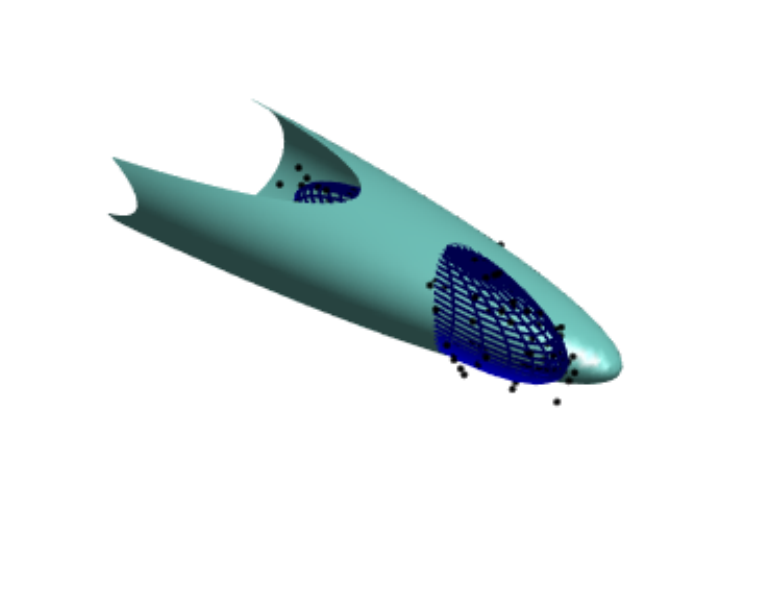}\\
			\includegraphics[width=1\textwidth]{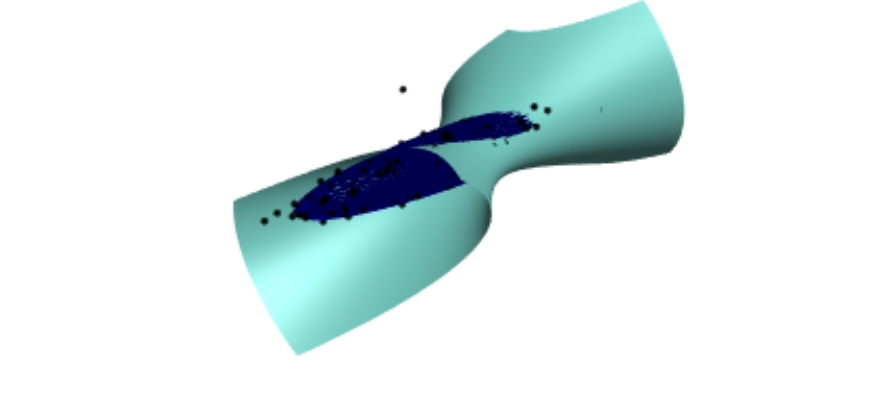}\\
			\includegraphics[width=1\textwidth]{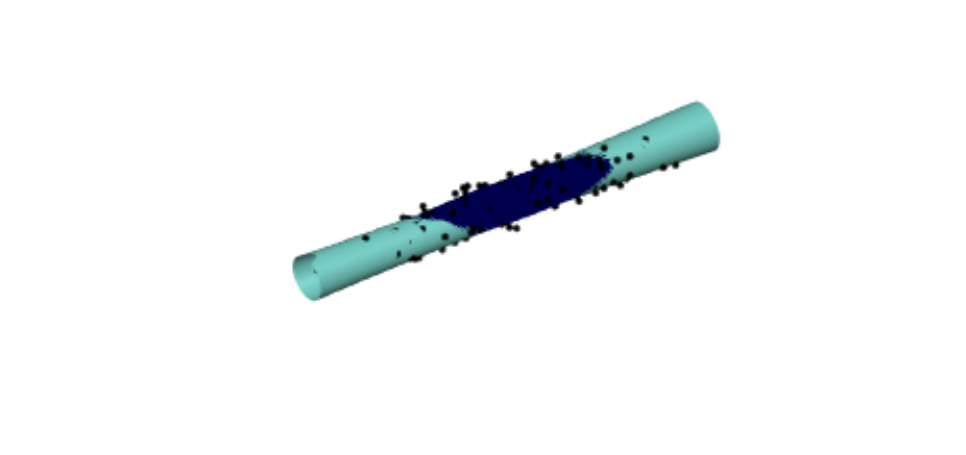}
		\end{minipage}
	}
	\subfigure[\tiny{Taubin~\cite{taubin1991estimation}}]{
		\begin{minipage}[b]{0.1\textwidth}
			\includegraphics[width=1\textwidth]{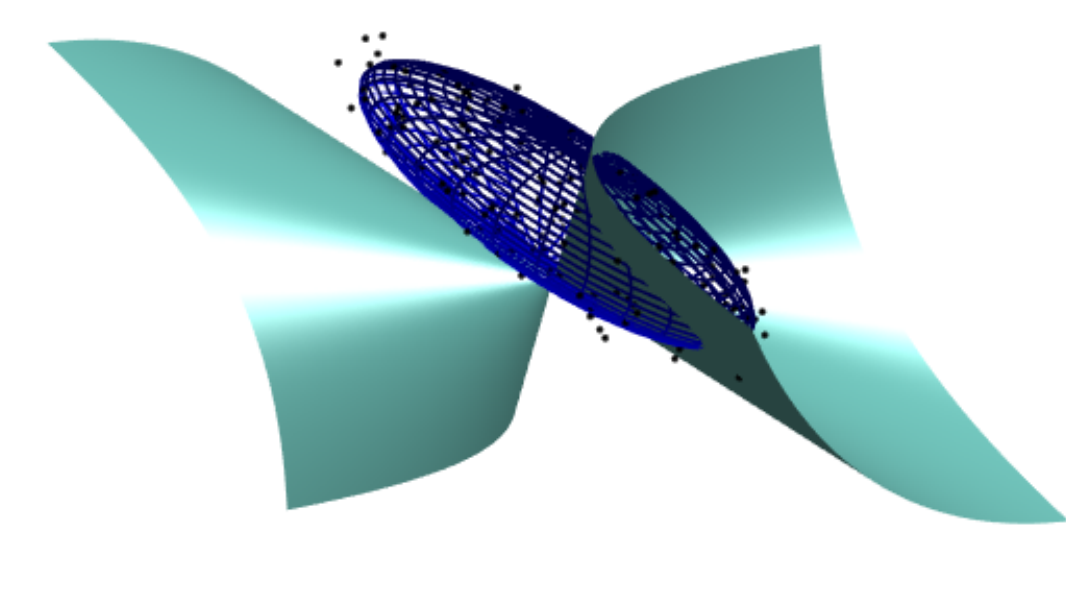}\\
			\includegraphics[width=1\textwidth]{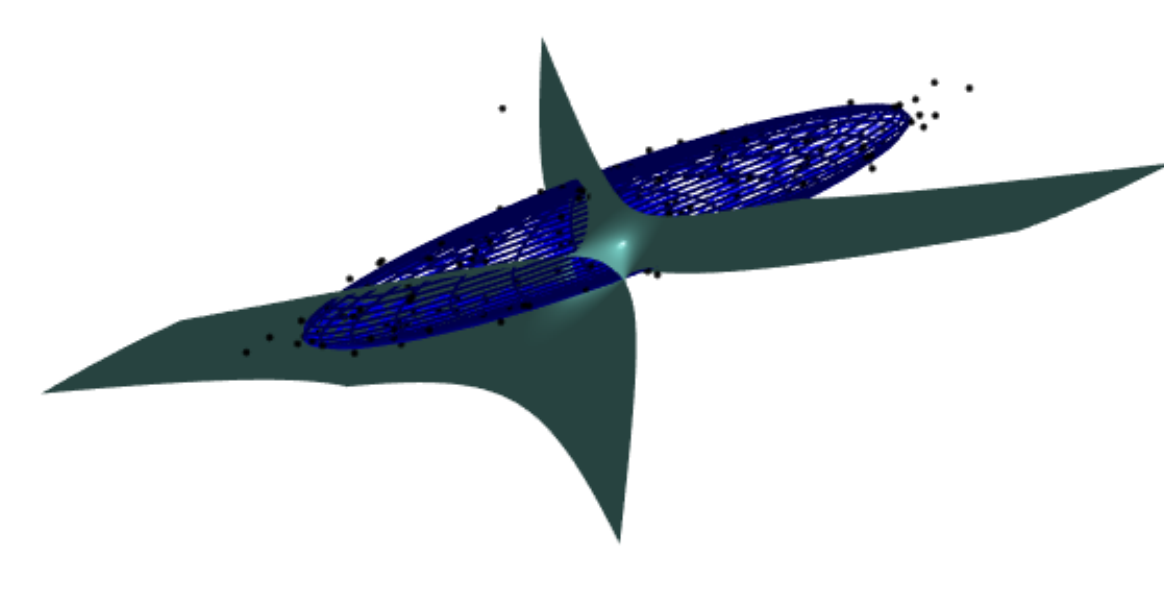}\\
			\includegraphics[width=1\textwidth]{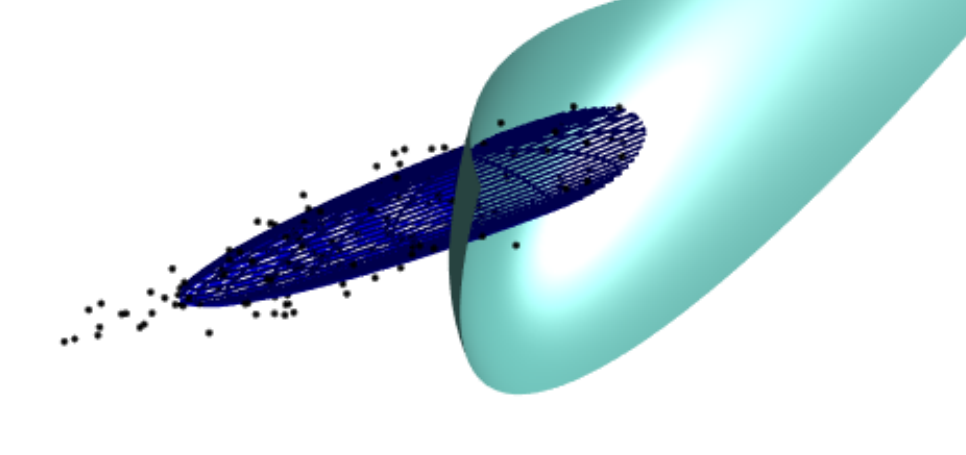}
		\end{minipage}
	}
	\subfigure[\tiny{GF~\cite{bektas2015least}}]{
		\begin{minipage}[b]{0.1\textwidth}
			\includegraphics[width=1\textwidth]{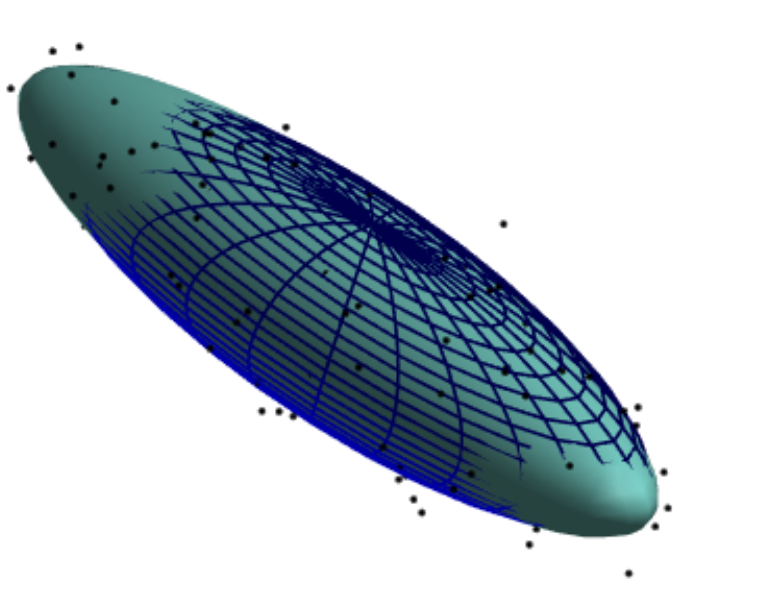}\\
			\includegraphics[width=1\textwidth]{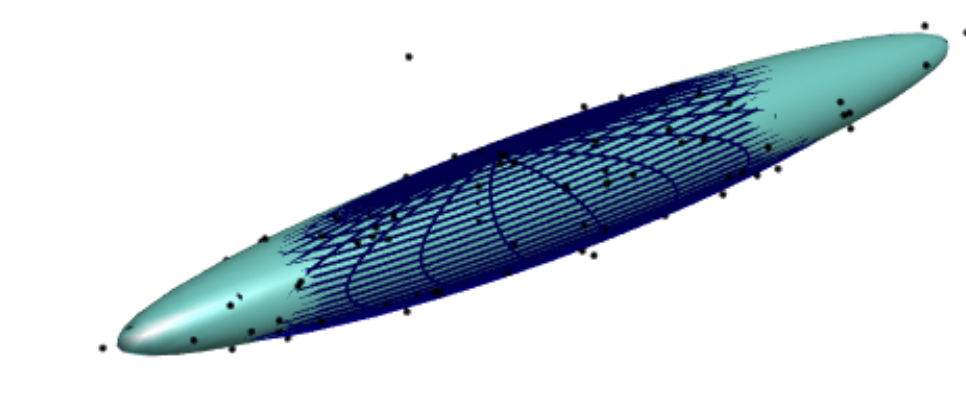}\\
			\includegraphics[width=1\textwidth]{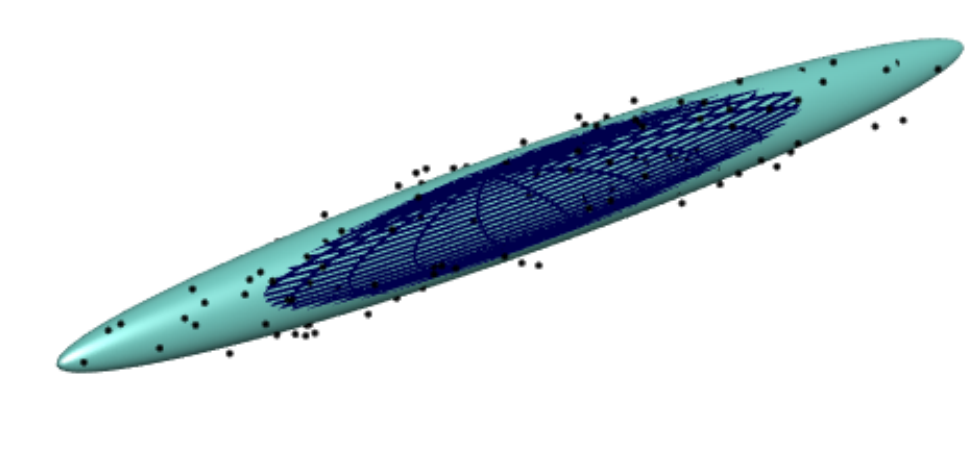}
		\end{minipage}
	}
	\subfigure[\tiny{RIX~\cite{lopez2017robust}}]{
		\begin{minipage}[b]{0.1\textwidth}
			\includegraphics[width=1\textwidth]{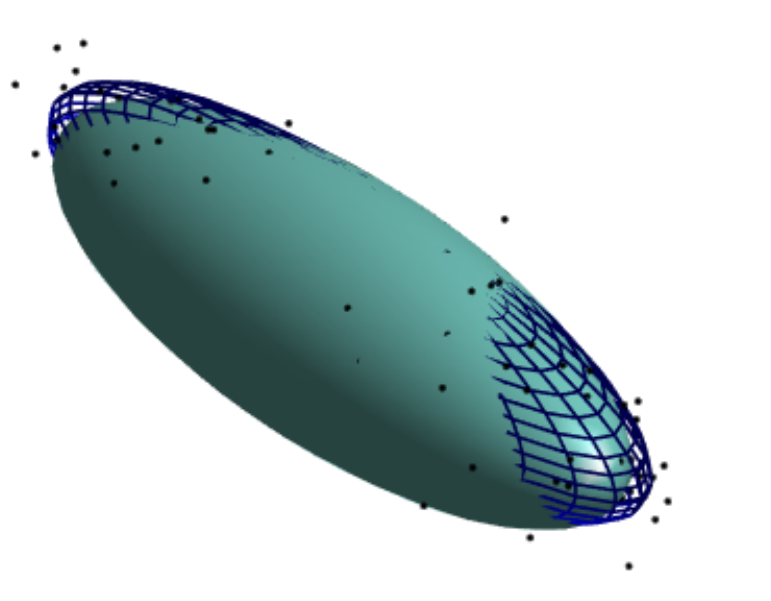}\\
			\includegraphics[width=1\textwidth]{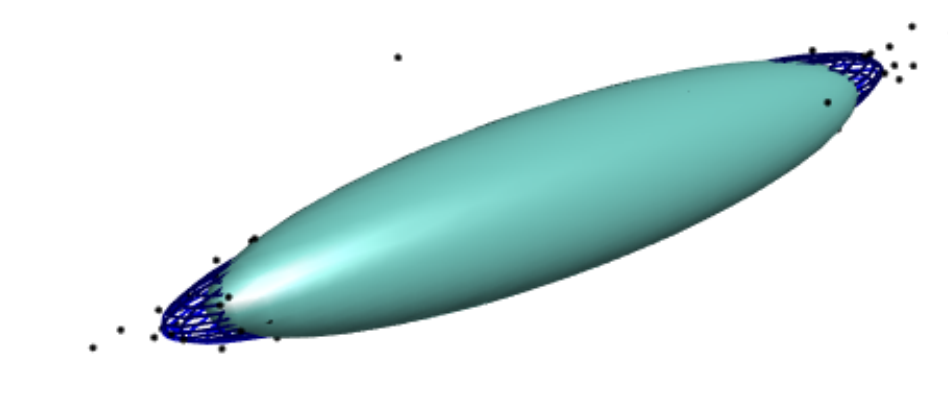}\\
			\includegraphics[width=1\textwidth]{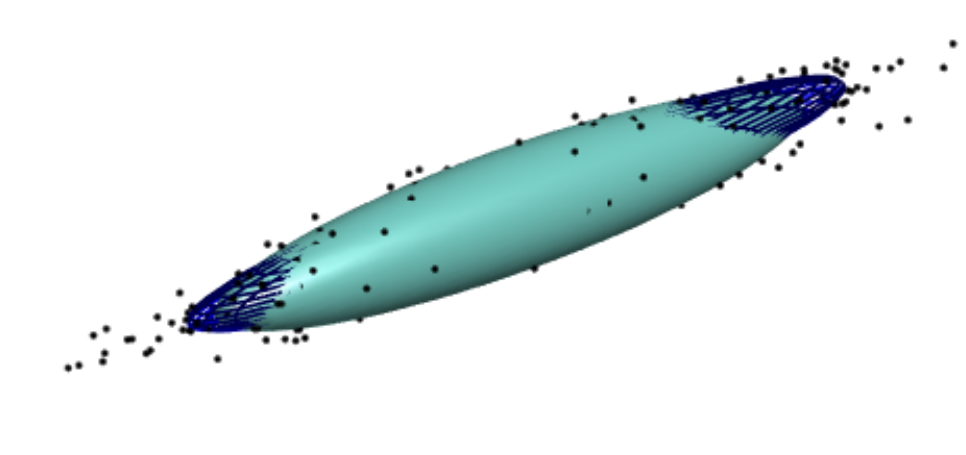}
		\end{minipage}
	}
	\subfigure[\tiny{Ours}]{
		\begin{minipage}[b]{0.1\textwidth}
			\includegraphics[width=1\textwidth]{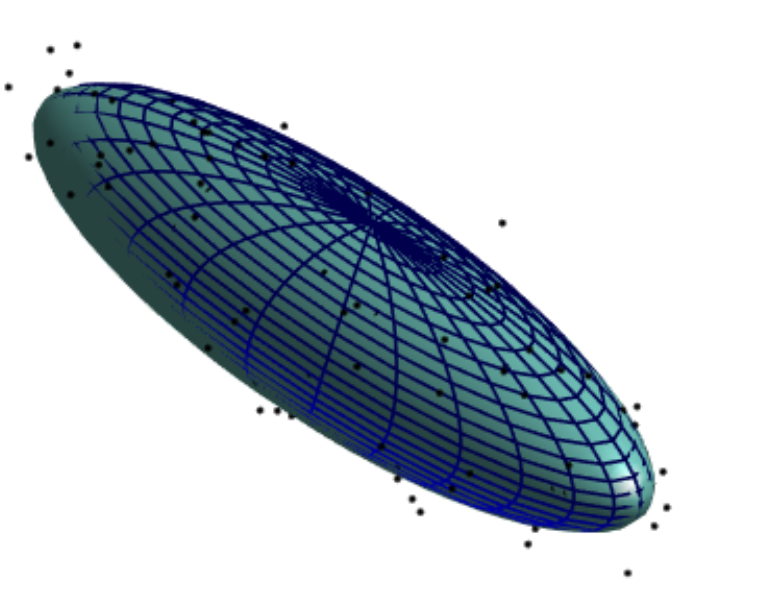}\\
			\includegraphics[width=1\textwidth]{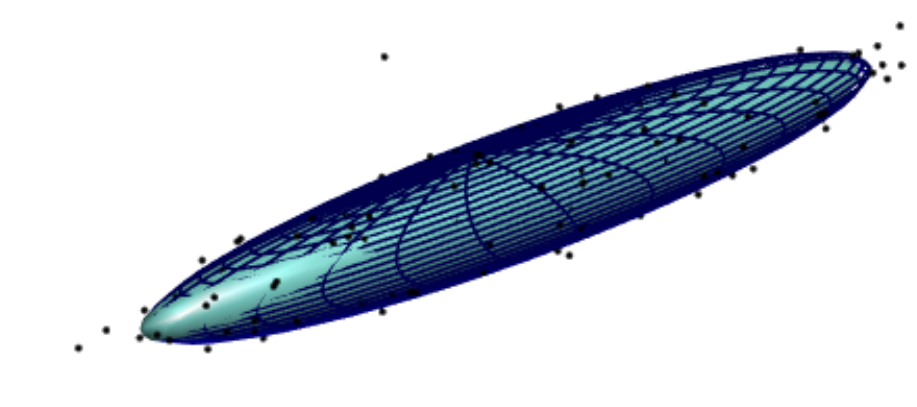}\\
			\includegraphics[width=1\textwidth]{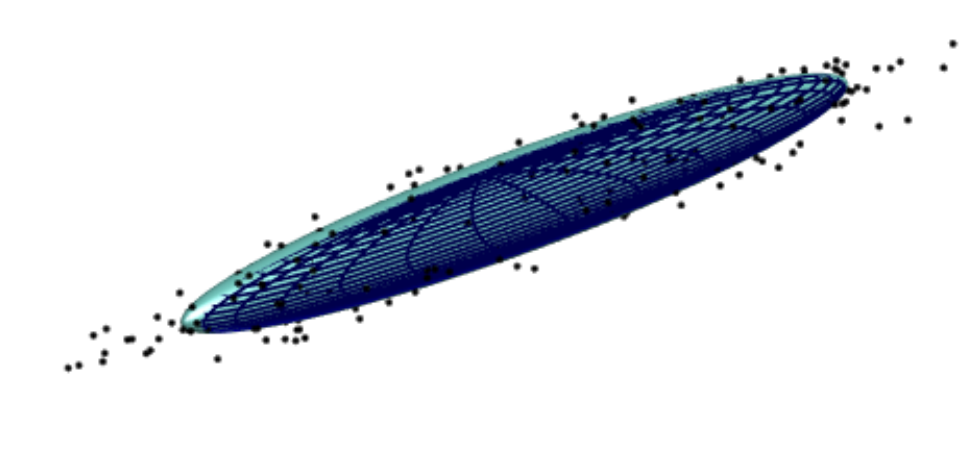}
		\end{minipage}
	}
	\vskip -0.1cm
	\caption{Influence of axis ratio $r_{ax}$ to different methods. Algebraic methods return significant errors, revealing their instability for thin or flat ellipsoids. Our method attains ellipsoid-specific fittings with the highest accuracy and is more robust against axis ratio.
	}
	\label{fig:axis}
	\vskip -0.3cm
\end{figure*}

\noin{Effect of outliers.}~Subsequently, we contaminate the ground truth data by a series of outliers from $5\%$ to $60\%$, along with zero-mean Gaussian noise, and  $\sigma=5\%$. Given that LS-based methods are susceptible to outliers,  \revise{we test the two robust methods and the iteratively re-weighted least-squares that use two M-estimators (robust kernels), such as Tukey~\cite{rousseeuw1991tutorial} and Huber~\cite{huber2004robust}. The results in the top right panel of Fig.~\ref{ellEx} show that RIX is relatively sensitive to outliers, especially when the outlier percentage exceeds $30\%$, which is consistent with the results reported by the authors~\cite{lopez2017robust}. M-estimators of Tukey and Huber have similar performance and are more robust than RIX, rooting from their weighting schemes for different residuals. Nevertheless, with outliers increasing at $60\%$, they also generate more fitting deviations.} In contrast, the proposed method works fairly well, and the deviations are kept quite low and stable, even when outliers rise up to $60\%$, demonstrating its high robustness. Comparison examples are presented in Fig.~\ref{fig:outlier_exam}.\\

\noin{Effect of the axis ratio.}~We also investigate the influence of the axis ratio $r_{ax}$ for ellipsoid fitting given that many existing ellipsoid-specific approaches require \emph{a prior} or have limitations for axis ratio. We randomly generate a set of ellipsoids with $r_{ax}$ from 1 to 5 and the statistical results are reported in the bottom right panel of Fig.~\ref{ellEx}. As observed, except our method, the others produce significant deviations with $r_{ax}$ increasing. Taubin and Koop are more sensitive to $r_{ax}$, MQF also showing its weakness. RIX exhibits noticeable shape deviations, whereas the proposed method achieves the highest accuracy for both metrics and keeps them greatly stable. Note that we have excluded non-ellipsoid fittings in the statistic. Three randomly generated ellipsoids with $r_{ax}=5, 8, 10$ (from top to bottom) and corresponding fittings are shown in Fig.~\ref{fig:axis}. \\

\begin{figure*}[t]
	\centering
	\includegraphics[width=0.37\textwidth]{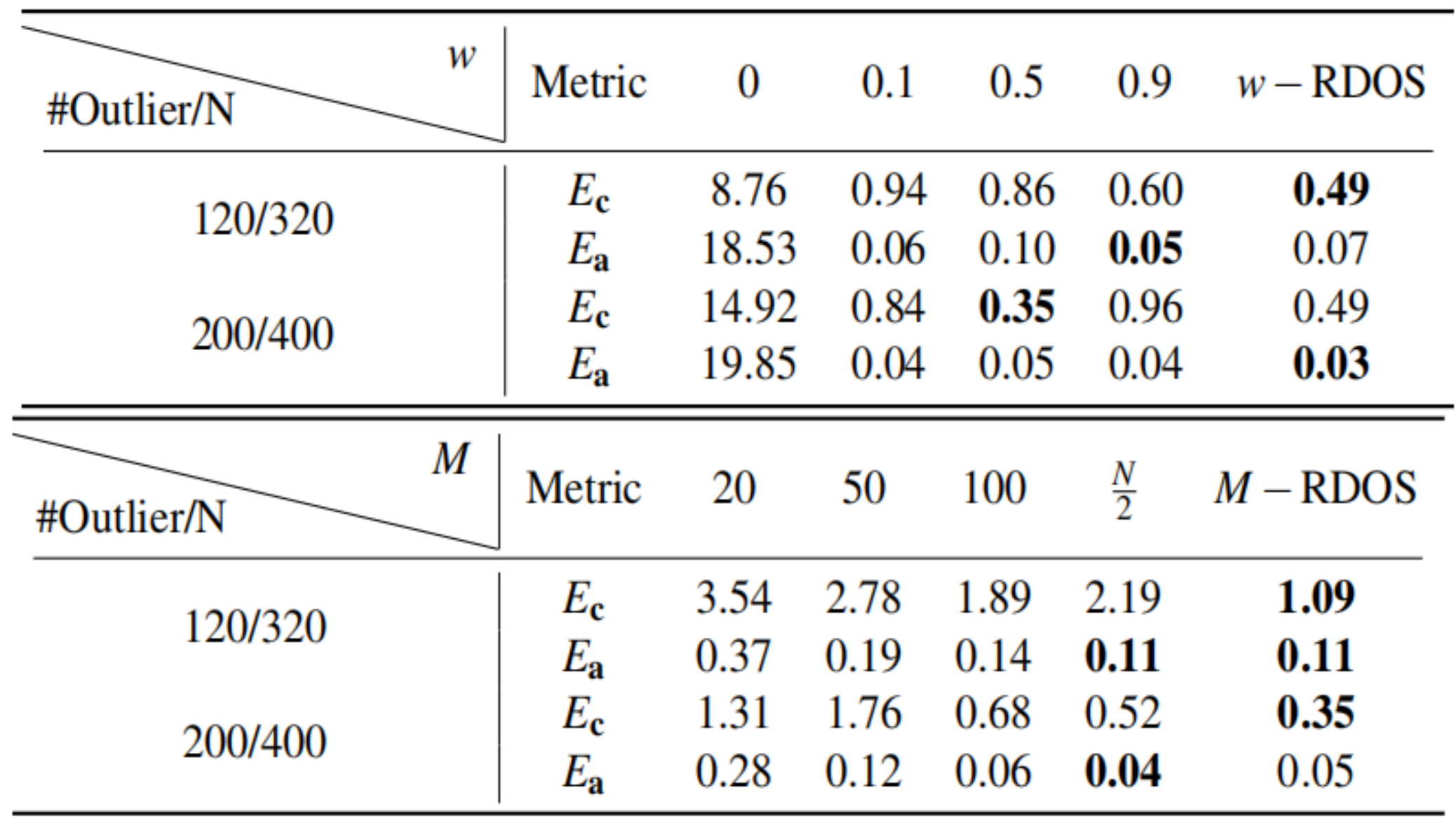}
	\includegraphics[width=0.6\textwidth]{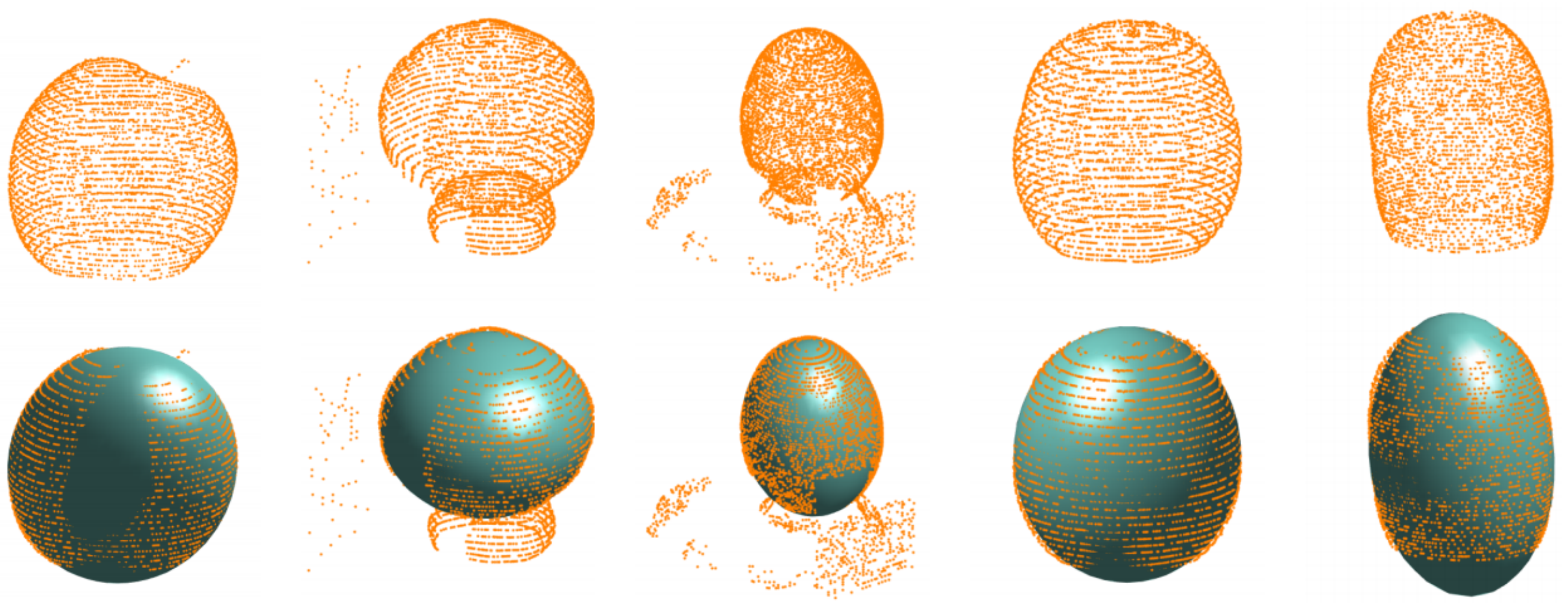}
	\caption{Left: Ablation study of RDOS for the initialization of weight $w$ (top) and the estimation of the inlier amount $M$ (bottom). \textbf{Bold font} indicates the top fitter; right: Our method attains successful fittings for point clouds with outliers and occlusion.}
	\vskip -0.5cm
	\label{fig:ablation}
\end{figure*}

\revise{
\noin{Ablation study of RDOS.}~RDOS is used to adaptively initialize the weight $w=\frac{\#\{\mathbf{x}_i|\RDOS(\mathbf{x}_i)>2\}}{N}$ that balances GMM and the uniform distribution because $w$ influences the performance and setting it manually may bring significant deviations. We conduct an ablation study by tuning different $w$ for two outlier-contaminated cases (120 and 200 outliers). Results in the top left panel of Fig.~\ref{fig:ablation}
 show that, compared with the random setting of $w \in \{0, 0.1, 0.5, 0.9\}$, $w-\RDOS$ ($w$ estimated by $\RDOS$) can provide more reasonable initialization, leading to an overall higher accuracy. Meanwhile, $M=\sum_{\mathbf{x}_i\in\mathbf{X}}\mathbbm{1}(\RDOS(\mathbf{x}_i)\leq 2)$ is taken to model sphere points. We also report the effect of $M$ for the previous two cases by fixing $w=0.375, 0.5$, respectively. Results in the bottom left panel of Fig.~\ref{fig:ablation} indicate that  more deviations will emerge if $M$ is much less than the number of inliers. On the contrary, $M-\RDOS$ ($M$ estimated by $\RDOS$) attains more satisfactory performance. Another simple choice is  let $M=N$ directly, but this choice will make $w=0$ in our method, resulting in significant errors. Despite we can tune $M$ by multiples of $N$ such as $\frac{N}{2}$, it may expand efforts to find a suitable value. Thus, we use adaptive $\RDOS$ for weight initialization and ellipsoid modeling, simultaneously.
} 

\noin{Real-world scanned point clouds.}~We next apply our method to 3D point clouds captured by a laser Picza scanner~\cite{lopez2017robust}. To boost fitting, we perform downsampling over the data (sampling rate $[0.2, 0.4]$), and there are $4,000\sim10,000$ points of each model. As shown in the right panel of Fig.~\ref{fig:ablation}, these point clouds bear evident occlusion or outliers that are usually \revise{disastrous} for LS-based methods. However, our fits exhibit acceptable results in the sense that the ellipsoid surfaces approximate the objects well, by which \revise{metrics} such as volume and direction can be estimated. \revise{Thereby, the proposed method in general is quite suited to densely sampled points attained by laser scanners or similar technologies.} 
\vskip -0.5cm

\vskip -0.3cm
\section{Discussion and Conclusion}
We have presented a robust and accurate method for ellipsoid-specific fitting in  noisy/outlier-\revise{contaminated} 3D scenes. We use GMM to  model the ellipsoid explicitly and cast it in \revise{an} MLE, which is effectively solved via the $\varepsilon$-accelerated EM framework. Furthermore, a uniform distribution is added to depress outliers, and all parameters are updated automatically. Comprehensive evaluations show that our method outperforms the compared ones by a large margin, especially for noisy, outlier-contaminated, ellipsoid-specific, and large axis ratio cases. 

\revise{
Given that our model is non-convex, EM may fall into local minima, but we scarcely see in previous experiments which may benefit from the proper initialization by RDOS. The number of mixture components in GMM depends on the measurement points, aiming to approximate arbitrary distribution. For efficiency, in future work, we can trim components based on the Gaussian bandwidth $h$ to let GMM adaptively model the ellipsoid. Furthermore, we can explore the use of a single analytic distribution over the surface of an ellipsoid for higher efficiency. Besides, we can replace Gaussian distribution by Student's t distribution~\cite{peel2000robust} to make the model more robust against noise with a heavy tail.
}

The proposed method can be generalized to fit other quadrics or conics, such as planes and cylinders, \revise{given the existence of} a parametric representation. We give a glance at other quadric fittings in the supplemental material. Additionally, we can boost the fitting accuracy by encapsulating more geometric features like normals and curvatures into the model.

\section*{Acknowledgments}
\vskip -0.3cm
This work is partially supported by the National Key Research and Development Program (2020YFB1708900), the National Natural Science Foundation of China (12022117, 61872354, 12171023, 62172415), the Beijing Natural Science Foundation (Z190004), the Open Research Fund Program of State Key Laboratory of Hydroscience and Engineering, Tsinghua University (sklhse-2020-D-07).

\bibliography{references}
\end{document}